\documentclass[12pt]{article}

\usepackage{newtxtext,newtxmath}
\usepackage{graphicx}

\usepackage[letterpaper,margin=1in]{geometry}

\renewenvironment{abstract}
	{\quotation}
	{\endquotation}

\date{}

\makeatletter
\renewcommand{\fnum@figure}{\textbf{Figure \thefigure}}
\renewcommand{\fnum@table}{\textbf{Table \thetable}}
\makeatother

\usepackage{scicite}

\usepackage{url}

\usepackage{setspace}

\usepackage{xcolor}
\def\scititle{
	Loop closure grasping: Topological transformations enable strong, gentle, and versatile grasps
}
\title{\bfseries \boldmath \scititle}

\author{
	Kentaro~Barhydt$^{1\dagger\ast}$,
	O.~Godson~Osele$^{2\dagger}$,
	Sreela~Kodali$^{2}$,
	Cosima~du~Pasquier$^{2}$,\and
	Chase~M.~Hartquist$^{1}$,
	H.~Harry~Asada$^{1}$,
	Allison~M.~Okamura$^{2}$\and
	\small$^{1}$Department of Mechanical Engineering, Massachusetts Institute of Technology, \\ 
        \small Cambridge, MA 02139, USA. \and
	\small$^{2}$Department of Mechanical Engineering, Stanford University, \\
	\small Stanford, CA 94305, USA. \and
	\small$^\ast$Corresponding author. Email: kbarhydt@mit.edu. $^\dagger$These authors contributed equally to this \and 
    \small work and are co-first authors.
}

\begin{document} 
% PI COMMENTS GENERAL NOTES

% Insert the title and author list
\maketitle

% Abstract, in bold
% There are strict length limits, and not all formats have abstracts.
% Consult the journal instructions to authors for details.
% Do not cite any references in the abstract.
\begin{abstract} \bfseries \boldmath
\noindent

Grasping mechanisms must both create and subsequently hold grasps that permit safe and effective object manipulation. 
Existing mechanisms address the different functional requirements of grasp creation and grasp holding using a single morphology, but have yet to achieve the simultaneous strength, gentleness, and versatility needed for many applications.
We present “loop closure grasping”, a class of robotic grasping that addresses these different functional requirements through topological transformations between open-loop and closed-loop morphologies. 
We formalize these morphologies for grasping, formulate the loop closure grasping method, and present principles and a design architecture that we implement using soft growing inflated beams, winches, and clamps.
The mechanisms’ initial open-loop topology enables versatile grasp creation via unencumbered tip movement, and closing the loop enables strong and gentle holding with effectively infinite bending compliance. 
Loop closure grasping circumvents the tradeoffs of single-morphology designs, enabling grasps involving historically challenging objects, environments, and configurations. 
\end{abstract}

\section*{Summary} 
Topological loop closure simultaneously enables versatile grasp formation and strong yet gentle retention of the grasp.

\section*{Introduction}

A successful grasp involves (1) creating a stable grasp and (2) subsequently holding that grasp in a manner that permits safe and effective object manipulation. 
These two stages of grasping have different functional and performance requirements. For the grasp creation stage, the grasping mechanism must articulate itself around the object to assume a desired, stable grasping configuration. 
Different objects and environments require different configurations, so the grasping mechanism must be versatile enough to achieve a wide variety of configurations. 
Improving versatility has long been a major motivation in grasping mechanism design research, especially soft mechanisms that can adapt to a wide variety of complex and unknown shapes \cite{samadikhoshkho_brief_2019, shintake_soft_2018, hughes_soft_2016, tai_state_2016, piazza_century_2019}. 
For the holding stage, the grasping mechanism must continuously resist destabilizing forces (e.g., gravitational, inertial, disturbance) during object manipulation to maintain the stable grasp without applying harmful contact pressures. 
This stage can be especially challenging for heavy yet fragile objects, which weigh significantly more than their allowable interaction forces, because of the conflict between the need for high forces to lift them and low force concentrations to avoid damage.

These different functional requirements for the different stages of grasping lead to distinct, and potentially conflicting, design needs. 
Existing grasping mechanisms employ a single static morphology \cite{samadikhoshkho_brief_2019, shintake_soft_2018, hughes_soft_2016, billard_trends_2019}, and have yet to achieve the simultaneously strong, gentle, and versatile performance needed for many high-impact applications involving heavy yet fragile objects. 
While previous designs have advantages for some of these performance aspects, none have been able to excel in all of them. 
For example, safely securing and lifting humans requires bulky manual tools and handler training due to their weight, fragility, pliability, articulation, and wide variance in shape and size. 
Handling humans is central to many critical applications such as elder care, care of people with physical disabilities, emergency medical response, search and rescue, physical rehabilitation, occupational therapy, and ergonomic support for manual labor \cite{dewit_fundamental_2013, 
% chang_impact_2010, 
feng_global_2019, lavender_designing_2007, 
% mortimer_risks_2011, 
furst_evaluation_2008, samdal_time_2019, penrose_occupational_1993, 
% behrman_physical_2007, 
norman_treadmill_1995, srivatsan_systematic_2024
% , hsiao_development_2009
}. 
Handling heavy yet fragile objects is also critical for many industry applications, such as agricultural harvesting, handling livestock, aircraft and ship manufacturing, civil and architectural engineering, construction, salvage recovery/excavation, and machinery installation \cite{bostelman_robocrane_1996, hoffman_precision_2020, glerum_stage_2024, kelechava_asme_2022, ibekwe_it_nodate, noauthor_crane_nodate, roy_maneuvering_2003}. 
However, these applications also require manual tools and trained human operators to harness and handle their respective payloads. 
Grippers designed to bear high loads exist, but can only grasp durable objects with no risk of damage, and cannot handle a wide variety of object sizes they can handle, the grasping configurations they can achieve, and the environments in which they can achieve those grasps (e.g., heavy manufacturing robots \cite{noauthor_bmw_2024, phillips_massive_2016}, previous patient lifting robots \cite{mukai_development_2010, loh_pneumatic_2014}).

We present a method of grasping that transforms the topology of the grasping mechanism’s morphology between open- and closed-loop to enable the creation and holding of simultaneously strong, gentle, and versatile grasps (Fig. 1). 
Our holistic method, which we call “loop closure grasping”, is centered specifically around this transformation in topology because of the fundamentally distinct and converse advantages of open- and closed-loop topologies for the creation and holding stages of the grasping process, respectively. 
By strategically enabling and employing transformations in the grasping mechanism’s morphology between both topologies, their advantages can be leveraged for the appropriate stage of grasping, and their disadvantages for the other stage can be bypassed. 
Thus, the tradeoffs of previous single-morphology designs can be circumvented. 
We additionally present a holistic grasping system design architecture and principles for enabling and leveraging this method. 
Prior examples of dynamically changing robot morphologies include reconfigurable robots and robots that adapt their morphology to their environment \cite{baines_multi-environment_2022, sihite_multi-modal_2023, sun_embedded_2023, nygaard_real-world_2021}. 
Unlike these previous works, loop closure grasping utilizes morphological changes for grasping.

\begin{figure} % Do NOT use \begin{figure*}
	\centering
	\includegraphics[trim={0.0cm 0.0cm 0.0cm 0.0cm}, clip, width=0.83\textwidth]{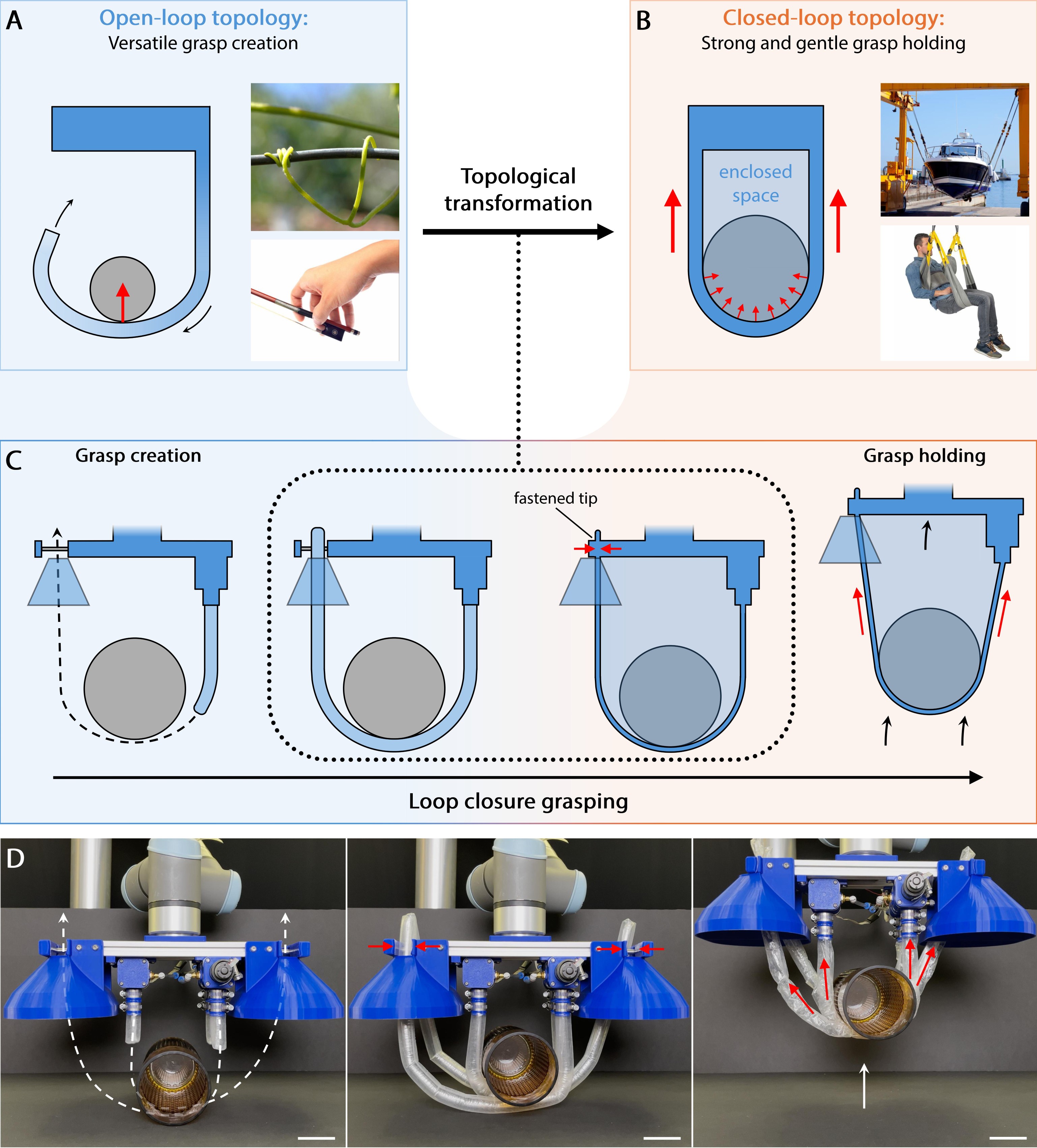} % for an image file named example_figure.*
	% Pick an appropriate width - in print, figures are usually one or two columns wide, which can
	% be approximated by 0.3\textwidth or 0.6\textwidth respectively. Use appropriate label sizes.

        % IMAGE SOURCES:
        % https://www.flickr.com/photos/blueridgekitties/7339966222 
        %   Google image result: creative commons license 
        % https://www.shutterstock.com/image-photo/close-elephant-trunk-holding-fruit-forage-1971207152
        %   Google image result: Commercial & other licenses 
        % https://www.pexels.com/photo/person-hand-with-chopsticks-eating-colorful-puzzles-9227216/
        %   Google image result: creative commons license 
        % https://www.medicalexpo.com/prod/horcher-medical-systems/product-68764-450122.html 
        %   No special license listed, CAN WE USE THIS? 
        % https://www.123rf.com/photo_14259959_crane-travelift-lifting-a-boat-on-blue-sky-day-in-balearic-islands.html
        %   Google image result: Commercial & other licenses ($12.30)
        % https://itoldya420.getarchive.net/amp/media/kangaroo-joey-699e52
        %   Google image result: creative commons license 
        % https://www.surehands.com/products/two-piece-hygiene-sling 
    {\linespread{1.0}
	\caption{\textbf{An overview of loop closure grasping.} The grasping process can be decomposed into two stages, grasp creation and grasp holding, connected by topological transformation. 
        \textbf{(A)} Grasping mechanisms with open-loop topologies are advantageous for versatile grasp creation, as illustrated by biological open-loop grasping mechanisms.
        \textbf{(B)} Grasping mechanisms with closed-loop topologies are advantageous for strong and gentle grasp holding, as illustrated by examples of closed-loop grasping mechanisms. 
        \textbf{(C)} Loop closure grasping process. Topological transformations enable mechanisms to transition their morphology from open-loop to closed-loop. By transforming its topology from open-loop to closed-loop between the grasp creation and holding stages, a grasping mechanism can leverage the advantages of the right topology at the right stage. 
        \textbf{(D)} A loop closure grasping system prototype grasping and lifting a glass vase. Scale bars, 5 cm.
        (A) Photo Credit: Leung Cho Pan, Alamy Ltd.; Photo Credit: oboltus, iStockphoto LP. (B) Photo Credit: tonobalaguer, Inmagine Lab Pte Ltd; Photo Credit: N/A, SureHands\textsuperscript{\textregistered} Lift \& Care Systems. Note that we have the licenses for, or permission to use these images and are entitled to use these materials by their associated owners.
    }}
	\label{fig:fig1} % give each figure a logical label name
\end{figure}

Open-loop mechanisms are advantageous for creating versatile grasps, as their tips can be moved freely around the object and environment to navigate the mechanism into a desired grasping configuration. 
Nearly all past grasping mechanisms have followed this open-loop paradigm, which has resulted in advances in versatile grasp creation \cite{samadikhoshkho_brief_2019, shintake_soft_2018, hughes_soft_2016, billard_trends_2019, cairnes_overview_2023, hernandez_current_2023, tai_state_2016, 
% bicchi_robotic_2000,
prattichizzo_grasping_2016, becker_active_2022, brown_universal_2010, piazza_century_2019, aljaber_soft_2023, shintake_versatile_2016, qu_advanced_2024, he_grasping_2025
% , crooks_fin_2016
}. 

However, current open-loop grasping mechanisms still cannot effectively hold heavy yet fragile objects on the order of heavy industry payloads, humans, etc. \cite{samadikhoshkho_brief_2019, shintake_soft_2018, hughes_soft_2016, billard_trends_2019, cairnes_overview_2023, hernandez_current_2023, tai_state_2016, 
% bicchi_robotic_2000, 
prattichizzo_grasping_2016, li_fluid-driven_2017, cacucciolo_delicate_2019, loh_pneumatic_2014}. 
Holding strong grasps with open-loop mechanisms requires significant flexural rigidity to resist destabilizing forces (see Results section for the generalized principle), limiting the compliance of the grasp. 
Compliance enables gentle interactions by passively distributing applied forces (as is well established in soft robotics literature \cite{shintake_soft_2018, hughes_soft_2016, majidi_soft_2014, laschi_soft_2016, cianchetti_biomedical_2018, yasa_overview_2023}). 
Thus, a tradeoff exists between how strong and gentle the grasp can be. 

Conversely, closed-loop mechanisms are advantageous for holding strong and gentle grasps, but are limited in creating versatile grasps. 
Flexible closed-loop mechanisms that hold objects in a cradled suspended state can bear high loads while passively distributing them over a large contact area. 
These mechanisms have been used for millennia (e.g., ropes, slings) \cite{magazine_new_nodate, noauthor_sling_2024, langley_mobile_2020, suddendorf_its_2020} to safely hold heavy yet fragile objects \cite{glerum_stage_2024, kelechava_asme_2022, noauthor_crane_nodate, roy_maneuvering_2003, ibekwe_it_nodate, noauthor_hoisting_2002}. 
Their load capacity is dependent only on their tensile strength (see Results section for generalized principle). 
Thus, they can hold objects with theoretically “infinite bending compliance” (zero flexural rigidity), taking one of the strongest motivations for soft robotics (compliance \cite{majidi_soft_2014, laschi_soft_2016, cianchetti_biomedical_2018, yasa_overview_2023}) to one of its theoretical extreme.

% Examples of closed-loop grasping mechanisms being leveraged for safely securing heavy yet fragile objects can be found in both nature and industry. 
% Some examples from nature include marsupials’ pouches to securely cradle their young, suspensory ligaments in human eye sockets that gently hold the eye in place (29), and rib cages of sloths whose organs are gently cradled inside their skeleton when they passively hang from tree limbs (30). 
% Engineered examples in industry include slings for patient transfers, stretcher harnesses for search and rescue, and loading straps for heavy industry transport and manufacturing (31–35). 

However, permanently closed-loop grasping mechanisms are limiting for versatile grasp creation. 
In two dimensions, an object cannot physically enter a closed loop. 
In three dimensions, the entire loop must be maneuvered around the object at once due to lacking a free tip (unlike open-loop continuum and hyper-articulated robots that only need to navigate their tips). 
Thus, available paths through which it can navigate to achieve stable grasps are limited. 
Most current uses of closed-loop mechanisms to hold objects require external work to securely harness them \cite{glerum_stage_2024, kelechava_asme_2022, roy_maneuvering_2003, noauthor_crane_nodate, ibekwe_it_nodate, noauthor_hoisting_2002} (e.g., a human manually wrapping a sling around a heavy industry payload for lifting \cite{dewit_fundamental_2013}), and the few that can autonomously create grasps are limited in their versatility to specific objects in controlled environments \cite{kang_grasping_2023, manes_soft_2024}. 

By enabling the mechanism to dynamically transform between open- and closed-loop topologies, loop closure grasping can create and hold grasps that leverage the advantages of the right topology at the right stage. 
This eliminates the tradeoffs of single static morphology designs, marking a significant departure from existing paradigms. 

To fully enable the loop closure grasping method, a grasping mechanism design must not only enable its open-loop form to fasten its tip to its base to transform into a closed loop, but its open-loop form must enable the respective advantages for versatile grasp creation, and the closed-loop form must enable the respective advantages for simultaneously strong and gentle grasp holding. 
However, nearly all open-loop mechanisms with the tensile strength and bending compliance necessary to leverage the closed-loop topology do not have the articulation capabilities necessary to create versatile grasps, because they require external work to bring its tip to its base (e.g., a caregiver must attach the ends of a sling to a base frame to close the loop around a patient’s body). 
Additionally, open paths to create the desired stable grasping configuration around the object are often obstructed. 
For example, objects are almost always on a resting surface, which can prevent the mechanism from maneuvering underneath it to form a closed loop that can lift it from below. 
Other objects in the environment can also impede the necessary paths. 
These obstacles can significantly limit the variety of objects that can be grasped and configurations that can be achieved, especially in environments that are not specifically constructed to make the object accessible (e.g., not a factory or lab environment \cite{barhydt_high-strength_2023, kang_grasping_2023, manes_soft_2024}).
To address these challenges for realizing loop closure grasping, we utilize soft growing inflated beam robots, or vine robots \cite{hawkes_soft_2017, blumenschein_design_2020}, as both an open-loop mechanism to navigate around the object through highly constrained environments to create versatile grasps, and subsequently as a closed-loop mechanism to hold simultaneously strong and gentle grasps. 

The primary contribution of this work is the concept and design of loop closure grasping to enable simultaneously strong, gentle, and versatile grasps by leveraging the separate advantages of topologically open- and closed-loop morphologies. We present key principles that elucidate the fundamental benefits of sequencing topologically open-loop and closed-loop morphologies for safe, secure, and versatile grasping, and an implementation of our architecture using inflated beam vine robots \cite{hawkes_soft_2017} to demonstrate this method of grasping. 
Our implementation consists of vine robots made from high-tensile strength materials with high bending compliance, pressurized bases that drive the vine robot growth, and tip-fastening mechanisms to close the loop. 
We demonstrate and characterize the ability to create versatile grasps and apply strong and gentle pulling and holding forces to grasp and lift a 6.8 kg kettlebell weight buried in a cluttered bin of parts, grasp via weaving an enclosure around a ball, grasp and lift a ring and a bucket with a handle via topologically interlocking Hopf links, grasp and pull containers from 3 m away, safely lift humans from a bed, lift and perform “in hand” manipulation with a cylinder, and grasp and lift a watermelon.

%%%%%%%%%%%%%%%%%%%%%%%%%%%%%%%%%%%%%%%%%%%%%%%%%%%%%%%%%%%%%%%%%%%%%%%%%%%%%%%%%%%%%%%%%%%%%%
\section*{Results}

\subsection*{Loop Closure Grasping Method for Creating and Holding Grasps }

% \subsubsection*{Intro sentence to set up discussion of task/problem/model formalization before defining the actual method}

% The loop closure grasping method circumvents the tradeoffs of single-morphology gripper design paradigms for simultaneously strong, gentle, and versatile grasping by transforming the topology of the grasping mechanism to adapt to the distinct functional requirements of the different stages of grasping. 
To develop this method, we first decompose the grasping process into its functional stages and associated requirements, and formalize a simple model of grasping mechanism morphology to define its topological classifications in the context of grasping.

% \subsubsection*{Formalization of grasping task}

We decompose the process of grasping into the two stages of creating and subsequently holding the grasp. 
The functional requirements for these stages, respectively, are (1) the grasping mechanism must articulate itself around the object to assume a desired, stable grasping configuration, and (2) the grasping mechanism must continuously resist destabilizing forces during object manipulation to maintain the stable grasp without applying harmful contact forces.
The key performance factor for the grasp creation stage that we consider is grasp versatility, which we define in terms of the variety and scale of objects it can grasp, the variety of stable grasping configurations it can deploy, and the variety of environments within which it can grasp these objects.
The key performance factors for the second stage are how securely and gently it can hold heavy yet fragile objects. 
We define a heavy yet fragile object as one with a significantly high weight relative to its allowable interaction forces. 
See Supplementary Materials S1 for more detailed definitions.

% \subsubsection*{Topological model of grasping mechanisms}

\begin{figure} % Do NOT use \begin{figure*}
	\centering
	\includegraphics[width=1.0\textwidth]{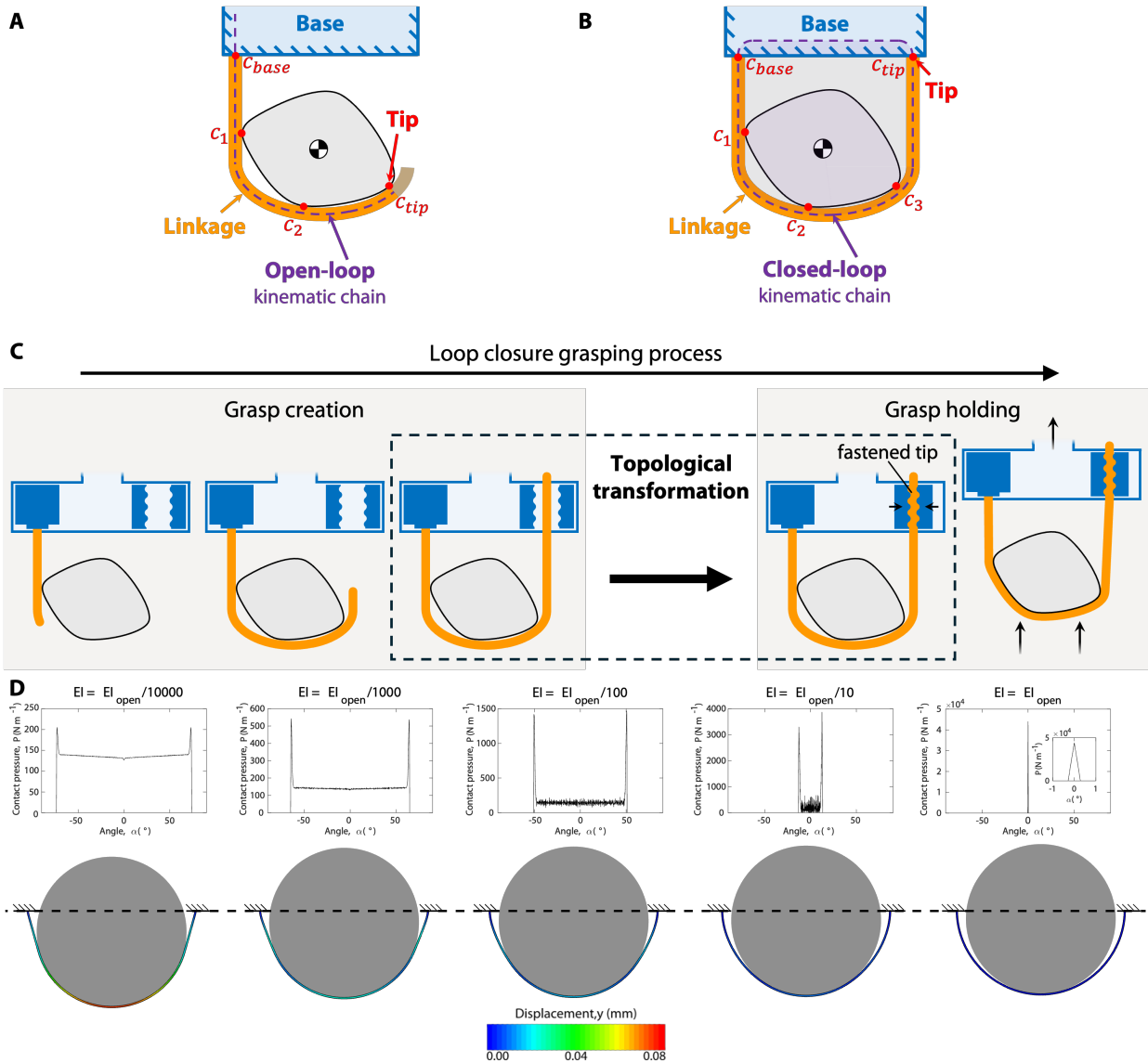} % for an image file named example_figure.*
	% Pick an appropriate width - in print, figures are usually one or two columns wide, which can
	% be approximated by 0.3\textwidth or 0.6\textwidth respectively. Use appropriate label sizes.

	% Captions go below figures
     {\linespread{1.0}
	\caption{
    \textbf{Topological model of grasping mechanisms and the loop closure grasping method.} 
    \textbf{(A-B)} Defining components and features of an open-loop and a closed-loop grasping mechanism, respectively. $c$ denotes a contact point.
    \textbf{(C)} Formal loop closure grasping method, sequenced in its three steps: open-loop grasp creation, topological transformation, and closed-loop grasp holding. \textbf{(D)} Pressure distributions between a circular object and five different closed-loop mechanisms with different flexural rigidities. 
    }}
	\label{fig:fig2} % give each figure a logical label name
\end{figure}

We define the open-loop and closed-loop morphological classifications for grasping mechanisms based on their topology relative to the object. 
We incorporate the object’s location into our definition because we are interested in how the grasping mechanism interacts with the object, not just the mechanism itself. 
As shown in Fig. 2A-B, the components of all grasping mechanisms can be categorized into three parts: the base, linkage, and tip. 
The grasping mechanism is grounded to the base (typically fixed to some manipulator). 
The linkage articulates/activates to make contacts with the object to constrain the object's position relative to the base. For our definition, we only care about the geometric path of the linkage as it connects the base to its points of contact with the object. 
Thus, for generality, we abstract the linkage as a single kinematic serial chain (discrete or continuous) starting at the base and ending at the most distal contact point. 
We define this most distal contact point as the tip. We consider grasping devices with multiple serial kinematic chains as having multiple grasping mechanisms used together.
For example, the grasping mechanism in human hands is the finger, of which five are grounded to the palm (the base). 
Similar to how multiple open-loop grasping mechanisms can be used in tandem to create a fully caged grasp in three-dimensional space \cite{makita_3d_2008}, multiple closed-loop mechanisms can accomplish this as well. 
See Supplemental Materials for additional definitions regarding branching kinematic chains and mechanisms with geometries defined by two-dimensional manifolds. 

We consider a grasping mechanism to have a topologically open-loop morphology when its tip is not grounded (Fig.2A). 
Thus, the tip is defined by the mechanism’s most distal contact point, and its position relative to the base is determined by the pose of the mechanism. 
We consider a grasping mechanism to have a topologically closed-loop morphology (Fig.2B) if it satisfies the following two criteria: (1) the tip is grounded to the same system as the base, and (2) the object is inside of the loop created by the mechanism. 
See Supplementary Materials and Fig.S1 for additional descriptions and arguments. 
% Similar to how multiple open-loop grasping mechanisms can be used in tandem to create a fully caged grasp in three-dimensional space \cite{makita_3d_2008}, multiple closed-loop mechanisms can be used to accomplish this as well. 

We now identify the key advantages and disadvantages of each topology for the different stages of grasping to inform how they can both be strategically employed to circumvent the tradeoffs of previous single-morphology designs. 
For the grasp creation stage, open-loop mechanisms are advantageous over closed-loop mechanisms for achieving high grasp versatility because their tips are free to move around the object and environment into the desired grasping configuration. 
Given sufficient bending degrees-of-freedom (DOFs), they can theoretically navigate through any pathway wider than the mechanism’s cross-section, as all points along its length can follow the same path (as established in continuum and hyper-articulated robotics literature \cite{laschi_soft_2016, hawkes_soft_2017, russo_continuum_2023, dong_development_2017, burgner-kahrs_continuum_2015, del_dottore_growing_2024}). 
Conversely, the closed-loop topology limits grasp versatility because the tip is fixed. 
The entire closed-loop mechanism must be maneuvered around the object at once, limiting the available paths that it can navigate through to achieve the desired grasp configuration. 

For the grasp holding stage, closed-loop mechanisms are advantageous over open-loop mechanisms for strong yet gentle grasping primarily because they can hold objects with theoretically “infinite bending compliance” (i.e., zero flexural rigidity) and passively distribute forces over a large contact area. 
Compliance enables gentle interactions with fragile objects as well established in soft robotics literature \cite{shintake_soft_2018,hughes_soft_2016,laschi_soft_2016,majidi_soft_2014,cianchetti_biomedical_2018,yasa_overview_2023}. 
Open-loop mechanisms require some amount of flexural rigidity to stably hold objects (i.e. resist pulling/destabilizing forces), limiting the compliance of the grasp. 
Greater holding forces require greater rigidity, and thus, a tradeoff exists between how strong and gentle the grasp can be.
Closed-loop mechanisms do not require any flexural rigidity to hold objects; instead, the load can be borne entirely by their tensile strength. 
In Supplementary Materials S3, we prove that this fundamental difference between open-loop and closed-loop grasping mechanisms is a generalizable feature inherent to their topology, and that this difference exists for holding and pulling on objects across all possible mechanisms. 
By eliminating flexural rigidity entirely, the closed-loop mechanism’s compliance when conforming to the object becomes theoretically infinite, eliminating bending reaction forces and, for convex objects, maximizing contact area. 
As shown in Supplementary Materials and Fig. S3, we illustrate this benefit by simulating the contact pressure distributions of closed-loop and open-loop grasping mechanisms to compare their pressure distributions.
Thus, the load capacity of closed-loop grasping mechanisms can be scaled up by increasing only the tensile strength of the mechanism without needing to increase flexural rigidity, eliminating the tradeoff between strong and gentle grasp holding inherent to open-loop mechanisms. 
Infinite bending compliance, in contrast to the high compliance used in previous grasping/harnessing mechanisms \cite{barhydt_high-strength_2023, kang_grasping_2023}, is particularly advantageous because minimizing flexural rigidity in closed-loop suspensory mechanisms is always beneficial until it goes to zero. 
Grandgeorge et al. \cite{grandgeorge_elastic_2022} showed that pressure concentrations increase at the two touch-down points on either side of the object as flexural rigidity increases. 
This is further shown in the simulation of contact pressure distributions of closed-loop mechanisms with increasing flexural rigidity in Fig. 2D.
Given that structures are generally much stronger/stiffer in tension than in bending, the mechanism’s thickness can also be kept low while maintaining high load capacities, and thus can be made longer/wider for an allowable mass/volume to increase contact area and further decrease contact pressures. 
Additionally, the closed-loop topology also enables significantly higher grasp stability. 
For planar grasps, the mechanism topologically surrounds the entire object in a closed loop, and thus the object can only leave the grasp if the loop breaks.
Similar to traditional multi-fingered grasping devices, multiple closed loops can be combined to create this closure for three-dimensional grasps \cite{kang_grasping_2023}.

% \subsubsection*{Formal definition of loop closure grasping method}

Building on these elucidations of the grasping process, topological mechanism classifications, and their respective advantages and disadvantages, we now present the loop closure grasping method. 
Loop closure grasping is enabled by transforming the topology between the grasp creation and holding stages by fastening the mechanisms’ tips to its base frame (or to each other). 
The method is illustrated in Fig. 2C.

The method is a sequence of three steps: (1) open-loop grasp creation, (2) topological transformation, and (3) closed-loop grasp holding. 
In the first step, the grasping mechanisms have an open-loop topology to perform the grasp creation stage. 
The mechanisms articulate around the object to assume the form for a desired stable closed-loop holding configuration. 
This includes positioning its tip relative to the base (or another mechanism’s tip) such that it can be fastened. 
The open-loop topology of the mechanism facilitates versatile grasp creation, enabling versatile configurations for a wide variety of objects and environments. 
In the second step, the grasping mechanism’s topology is transformed to a closed loop by fastening its tip to the base frame (or another mechanism’s tip) without changing its configuration. 
To preserve the desired grasp configuration facilitated by the mechanism’s previously open-loop topology, the fastening method cannot alter the configuration. 
In the final step, the grasping mechanisms have a closed-loop topology to perform the grasp holding stage. 
The mechanisms’ articulation becomes passive with minimal flexural rigidity to realize its “infinite” bending compliance.
The closed-loop mechanisms’ configurations (created during the first step) must stably hold the object in a cradled suspended state (or a tightly cinched state by retracting its length) when the base is manipulated to pull/lift the object. 
The closed-loop topology enables grasp holding with “infinite” bending compliance, facilitating strong and gentle holding of heavy yet fragile objects. 

To release the grasp, the reverse of the first two steps can be performed. 
The tips of the mechanisms can be unfastened to transform the topology back to open-loop, at which point they can articulate (or retract) to remove themselves away from the object.

\subsection*{System Design Architecture and Principles}

\begin{figure} % Do NOT use \begin{figure*}
	\centering
	\includegraphics[width=1.0\textwidth]{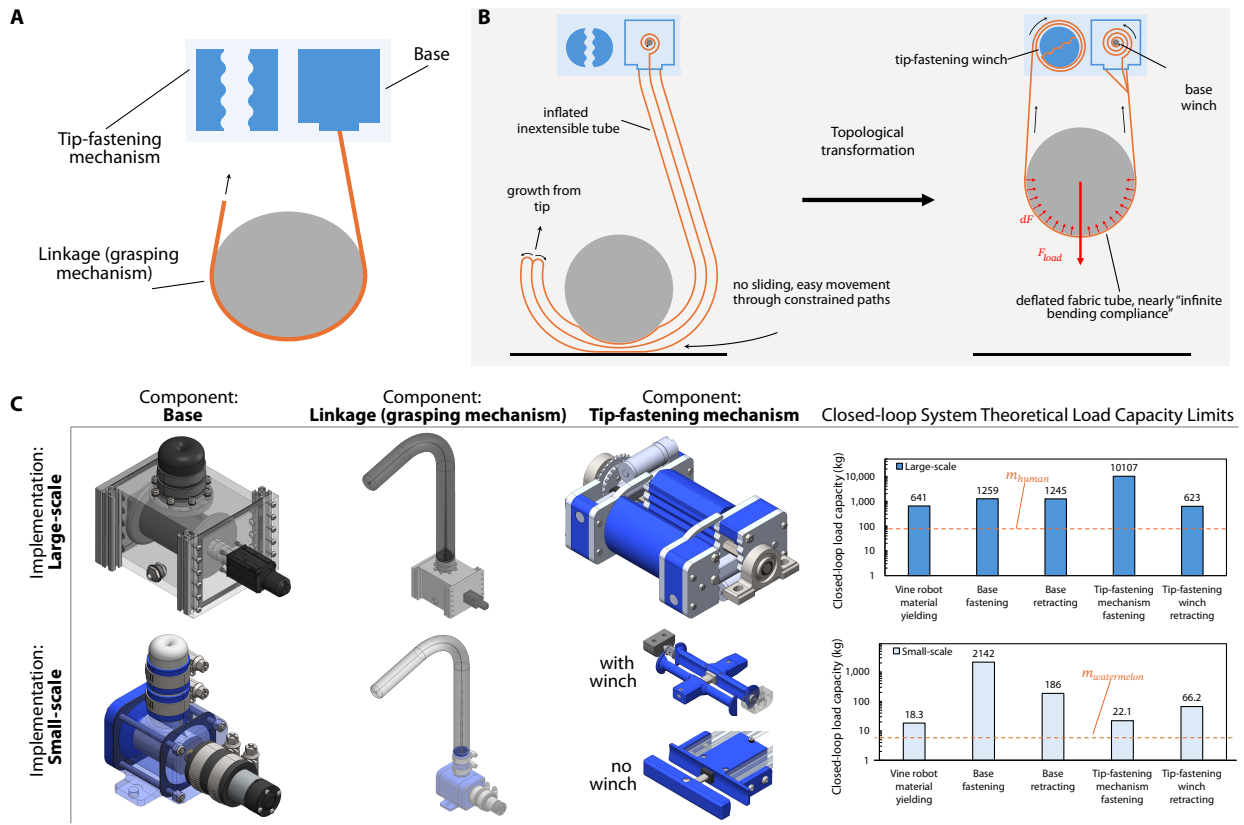} % for an image file named example_figure.*
	% Pick an appropriate width - in print, figures are usually one or two columns wide, which can
	% be approximated by 0.3\textwidth or 0.6\textwidth respectively. Use appropriate label sizes.

	% Captions go below figures
     {\linespread{1.0}
	\caption{
        \textbf{Loop closure grasping system design architecture and components design.} 
        \textbf{(A)} System design architecture with components labeled. 
        \textbf{(B)} Robotic implementation using soft growing inflated beams (i.e., vine robots), clamps, and winches. 
        Our implementations of these mechanisms satisfy the design principles for each component and the overall architecture to enable and realize the loop closure grasping method. 
        \textbf{(C)} Component designs for our large-scale and small-scale sets of system modules. Theoretical load capacity limits of the system due to vine robot material yielding, and the retraction force limits and fastening load capacities of the base and the tip. 
    }}
	\label{fig:fig3} % give each figure a logical label name
\end{figure}

Here we present a simple system design architecture and associated design principles for grasping systems that enable and realize the loop closure grasping method. 
This includes the system components, how they interrelate, and sets of principles for the components’ designs to satisfy the definitions, requirements, and goals of loop closure grasping.

% \subsubsection*{Model of grasping system and its components}

The system design architecture is illustrated in Fig. 3.
It consists of three components: (1) the base, (2) the grasping mechanism (i.e. the linkage), and (3) the tip-fastening mechanism. 
The proximal end of the linkage is mechanically grounded to the base.
Other components required to operate the linkage (e.g. actuators, transmissions, controllers) can be stored in the base as well, as is often done in traditional gripper designs. 
The tip-fastening mechanism is fixed to the same frame as the base, such that they are both manipulated as a single unit. 
The workspace of the linkage’s tip must also contain the tip-fastening mechanism such that they can engage to close the loop. 
Alternatively, tip-fastening mechanisms could be fixed to the tips of the bodies, enabling two bodies to fasten their tips together to create a single closed loop. 
The rest of our results focus on the former case.

% \subsubsection*{Component design principles}

The design principles that we propose for each of these components are as follows. 
The grasping mechanism (i.e. the linkage) must satisfy the requirements and goals of both the grasp creation and holding stages, as it is the component that actively creates the grasp configuration and interacts with the object while holding. 
Thus, it must be able to function as an open-loop and closed-loop grasping mechanism, and transform between them by fastening its tip to the same frame as its base. 

To create the grasp, the linkage must articulate itself into the desired grasping configuration without any additional external work when in its open-loop form. 
To enable versatile grasps, its open-loop form should have high bending DOFs (ideally continuum) and a large configuration/work space to navigate through complex paths to achieve a high variety of grasp configurations. 
It should also have a slender cross-section to enable navigation through constrained environments (e.g. through environmental obstacles) and obstacles (e.g. passing through a closed-loop in the object to create Hopf link \cite{rolfsen_knots_2003}).
While not fundamental to the architecture, additional beneficial functionalities include the ability to extend (so that the mechanism can follow the path of the tip, minimizing the space moves through to maneuver the tip), and the ability to make safe soft contact with the object while articulating around it (to enable configurations that start in contact with the object). 

To enable strong yet gentle grasp holding in its closed-loop form, the grasping mechanism should have a high passive tensile strength to bear heavy forces, and negligible flexural rigidity (i.e. infinite bending compliance) for gentle distribution of contact pressures. 
It should have high passive bending DOF (ideally continuum) and a large passive configuration/work space in its closed-loop form as well to enable high shape conformability and maximize contact area with the object (especially when convex). 
Additional beneficial, but not fundamental, functionalities include the ability to be retracted at its base and/or tip to enable active tightening of the loop to apply pulling and cinching forces, and the ability to pass between gaps smaller than its own width to navigate through especially constrained paths (e.g. get between objects and their resting surface).

The base and the tip-fastening mechanism do not actively create the grasping configuration or directly hold the object, and so their purposes are primarily to enable the grasping mechanism’s functionalities for loop closure grasping. 
The primary role of the base is to act as a ground frame for the grasping mechanism, and thus must be able to bear the same or greater loads than the grasping mechanism’s closed-loop tensile strength. 
The tip-fastening mechanism must be able to reversibly fasten the tip of the grasping mechanism to engage and release the closed-loop hold, and must also bear the same or greater loads than the grasping mechanism. 
The fastening mechanism design cannot impose any design requirements onto the grasping mechanism that would impede its performance and/or satisfaction of its respective design principles (e.g. requiring the grasping mechanism to have some fastening feature at its tip that would limit its ability to navigate through obstacles). 
An additional beneficial, but not fundamental, functionality for both the base and the tip-fastening mechanism is high-strength and reversible retraction of the grasping mechanism’s length to tighten the closed loop and/or apply pulling forces.

\subsection*{Grasping System Implementation with Soft Growing Inflated Beams}

% \subsubsection*{(Intro paragraph introducing that we implemented the system design architecture to develop two sets of component modules to enable the creation of loop closure grasping system prototypes for both small-scale and large-scale tasks.)}

To realize and demonstrate the loop closure grasping concept, we created a set of modular components to be used to create loop closure grasping systems (Fig. 3). 
% These modules were designed in accordance with our system design principles. 
We developed two sets of component modules, one for large-scale grasping and one for small-scale, with modules for the base, linkage or tip-fastening mechanism. 
Their different designs and constructions are described in detail in Methods and Materials and Supplemental Materials. 
A single set of these modules can be used to create a single-loop grasping system with one grasping mechanism. 
Multiple systems can be mounted to the same ground frame and used in tandem to create more sophisticated grasping systems.

% \subsubsection*{Component design}

Our design implementation for each component is as follows (Fig. 3): a high tensile strength pneumatically-driven soft growing robot (i.e., a vine robot \cite{hawkes_soft_2017}) as the linkage, a high-strength motorized base, and a high-force/torque tip-fastening winch device. 

We utilize vine robots as the grasping mechanism because their unique structure and mechanics fully satisfy, to an exceptional degree, all of our associated system design principles. 
Vine robots are long tendril-like soft inflated robotic structures that navigate their environments via pressure-driven “growth” from their tip, have an open-loop topology. 
They can also navigate through cluttered spaces without sliding friction (including between objects and their resting surface \cite{nakamura_soft_2018}) due to their tip-everting growth mechanism, grow to long lengths to wrap around large and/or far away objects, and grow into three-dimensional configurations against gravity due to their light weight \cite{blumenschein_helical_2018}.
Additionally, when deflated, vine robots can have a high tensile strength and nearly infinite bending compliance and conformability given proper material selection, precisely the characteristics needed to leverage the closed-loop topology for strong and gentle grasp holding.
While vine robots have previously only been utilized as open-loop mechanisms \cite{blumenschein_design_2020, coad_vine_2020}
% stroppa_shared-control_nodate, do_stiffness_2024}
, our systems incorporate tip-fastening mechanisms that secure the tips of vine robots to transform their topology to closed-loop, enabling their tensile strength and infinite bending compliance to be leveraged for grasp holding. 
% As shown in Fig.3D, our large-scale and small-scale vine robot grasping mechanisms have closed-loop tensile load capacities of 641 kg and 8.43 kg, respectively (details in Supplemental Materials). 
Thus, by navigating around the object, anchoring its tip onto its base via a tip-fastening mechanism, and deflating, vine robots can be used to create loop closure grasps.
% that fully leverage the advantages of both the open- and closed-loop topologies.

Like most bases for vine robots \cite{blumenschein_design_2020, coad_vine_2020}, our base design consists of a pressurized box that drives the vine robot growth, and an internal motorized winch to wind/unwind its inner material to control growth and apply pulling forces. 
This method of pulling via winding up the material is possible because of its “infinite bending compliance”, and enables work multiplication \cite{hawkes_engineered_2022} to lift heavy loads over long distances using relatively low-power actuators.
The load bearing inner material is permanently fastened to the winch, the strength of which is made sufficiently high (while maintaining a compact diameter) by ensuring that the inner material is always wrapped over some minimum angle so the fastening force is always magnified via capstan friction.
% As shown in Fig.3D, the fastening strength of our large-scale and small-scale base modules can enable an object weighing up to 1259 kg and 2142 kg to be lifted, respectively (details in Supplemental Materials). 

Our tip-fastening mechanism design consists of a clamp-winch device, as shown in Fig. 3C. 
For some of our small-scale demonstrations that did not require pulling from the tip, clamps without the winch function were designed and used for compactness. 
To reversibly fasten and pull on the vine robot, the tip-fastening winch closes its clamp after the tip has been inserted, and then rotates to wind up its length. 
Unlike many past vine robot related works \cite{jeong_tip_2020, kubler_multi-segment_2023, jitosho_passive_2023}, this clamping method enables the vine robot to be fastened directly, without requiring any tip-mounted devices that would impede its ability to navigate through highly-constrained paths into complex self-supported configurations. 
For its high reversible fastening strength, capstan friction from wrapping around the winch cannot be utilized here because the wrapping angle is zero when the vine robot is first fastened. 
However, capstan friction is still utilized through a wave pattern implemented in the clamp surface design that wraps the vine robot around a series of circular curved segments, as shown in Fig. S5. 
The holding force for a segment of the vine robot along one curve is determined by the load capacity of the segment along the previous curve.
Thus, the loading force is magnified at an exponential rate over the total wrapping angle across the series of curves, just like in the standard Euler–Eytelwein formula \cite{euler_remarque_1762}. 
In Supplementary Materials, we show that the load capacity of the wave-patterned clamp can be conservatively estimated as:
\begin{equation}
	T_{load} = \mu F_{clamp} e^{\mu (n+1) \theta_c}
	\label{eq:eq1}
\end{equation}
% As shown in Fig.3D, the fastening strength of our large-scale and small-scale tip-fastening mechanism modules can enable an object weighing up to 10107 kg and 12.1 kg to be lifted, respectively (details in Supplemental Materials). 
where $\mu$ is the coefficient of friction between the vine robot inner material and itself (assuming the load is only on the inner material, which is the case when the vine robot length is also retracted by the base), $F_{clamp}$ is the clamping force, $n$ is the effective number of curve segments on one of the clamp surfaces, and $\theta_c$ is their central angle.

The pulling strength of both the base and the tip-fastening winch depends on the gear ratio of their transmission and the maximum radius of their winch, including the thickness of the wound material. 
The retraction mechanism in both devices is simply the winch spooling up the length of the vine robot. 
% As shown in Fig.3D, the pulling strengths of our large-scale and small-scale base and tip-fastening mechanism modules enable an object weighing up to 1245 kg, 186 kg, 623 kg, and 66.2 kg to be lifted, respectively (details in Supplemental Materials). 

The strength of the overall grasp is determined by the weakest of all of these factors. 
The theoretical limits that each component imposes on the system, assuming it is the bottleneck, are shown in Fig.3D (details in Supplemental Materials).
Currently, the bottleneck factors for the large-scale and small-scale systems’ load capacities are the tip-fastening winch retraction force limit and vine robot tensile strength, yielding a 623 kg and 8.43 kg capacity for a single closed-loop vine robot mechanism, respectively. 
The large-scale load capacity bottleneck could be increased while keeping the motor torque capacity the same by reducing the diameter of the base winch or increasing the transmission’s gear reduction. 
The small-scale bottleneck could be increased by changing the membrane material or thickness.

\subsection*{Grasp Versatility}

% \subsubsection*{Versatility enables by initial open-loop morphology}

To demonstrate the grasp versatility enabled by utilizing an initially open-loop morphology, we performed three different grasps: (1) grasping a ball with four vine robots in a woven configuration, (2) grasping a ring and a bucket handle via interlocking closed loops with a single vine robot, and (3) grasping a kettlebell weight in a cluttered environment. 

\begin{figure} % Do NOT use \begin{figure*}
	\centering
	\includegraphics[width=\textwidth]{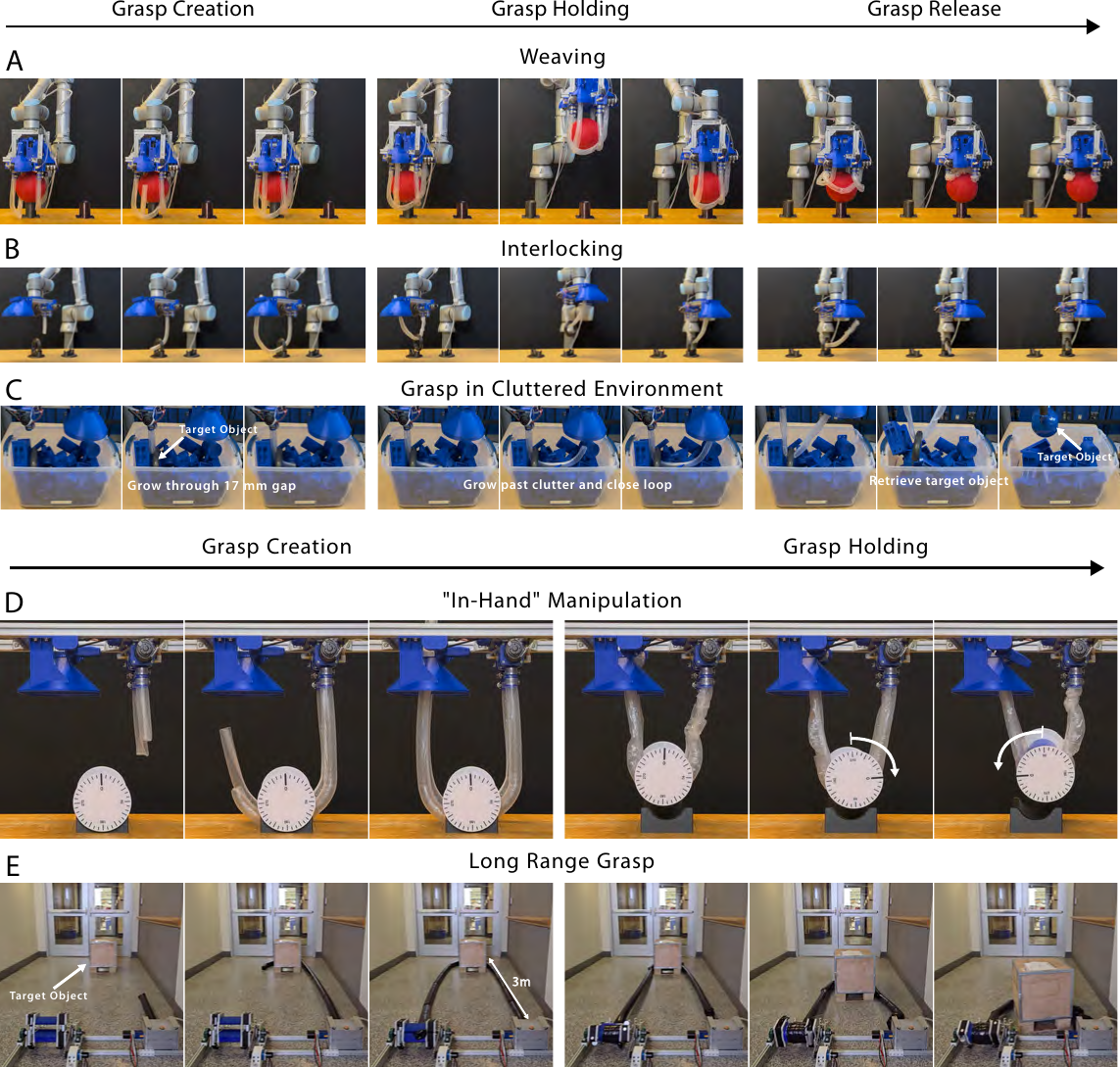} % for an image file named example_figure.*
	% Pick an appropriate width - in print, figures are usually one or two columns wide, which can
	% be approximated by 0.3\textwidth or 0.6\textwidth respectively. Use appropriate label sizes.

	% Captions go below figures
    {\linespread{1.0}
	\caption{
        \textbf{Grasp versatility demonstrations.} \textbf{(A)} Weaving around an object to secure it \textbf{(B)} Creating Hopf link by interlocking with ring \textbf{(C)}  “In hand” manipulation of a grasped object. \textbf{(D)} Grasping object in a cluttered environment \textbf{(E)} Grasping an object rested 3 m away from grasping mechanism.}}
	\label{fig:fig4} % give each figure a logical label name
\end{figure}

To demonstrate the ability to create grasps with different configurations, we used four vine robots to create a woven configuration around a ball to grasp and lift it. 
As previously demonstrated in \cite{kang_grasping_2023}, the entanglement of the woven warp and weft threads create a mechanically stable three-dimensional enclosure to grasp the object, which they achieved by twisting flexible closed loops such that they closed down onto the object from an initially sprawled configuration. 
In our demonstration, the vine robots were grown directly into the closed configuration, as shown in Fig. 4A and Movie S1, allowing for grasps in constrained environments. 
Using this configuration, we grasped and lifted a 178 mm diameter rubber ball, which would otherwise be more difficult to stably lift with an unwoven configuration. 
The grasp was successful and withstood inertial disturbances from manipulation. 
To illustrate the ability to release the grasp, the tips of the vine robots were subsequently released and their lengths were retracted back into their bases, placing the ball in a new location. 

To demonstrate the ability to grasp objects with different topologies, we used both our small-scale and large-scale grasping system prototypes to create interlocked Hopf links \cite{adams_knot_2004} between a single vine robot objects with permanently closed loops, namely a ring and the handle of the bucket. 
As shown in Fig. 4B and Movie S2, by using an initially open-loop mechanism to navigate through the object before closing the loop, we create grasps that would be topologically impossible with a permanently open- or closed-loop grasping mechanism. 
Assuming neither loop breaks, this grasp is theoretically infinitely stable to disturbances. 
To the best of the authors’ knowledge, this is the first demonstration of robotic grasping with this topological configuration. 

To demonstrate the ability to grasp objects in difficult environments, we grasped a kettlebell in a cluttered environment (e.g., a bin filled with loose parts) and grasped a glass vase resting on a table. 
Creating versatile grasps with permanently closed-loop grasping mechanisms can be challenging when obstacles exist in the environment because the entire closed loop must maneuver around the object at once (e.g. \cite{kang_grasping_2023, manes_soft_2024}, lassos). 
Conversely, open-loop mechanisms can theoretically navigate through any path wider than its cross-section \cite{hawkes_soft_2017}.
Further, vine robots can even pass through paths narrower than its own width. 
As shown in Fig. 4D and Movie S3, we used a single vine robot to successfully grasp and lift the kettlebell surrounded by various objects of different shapes and sizes. 
The smallest gap that the vine robot had to navigate through was approximately 17 mm wide, roughly 35\% smaller than the 26mm diameter of the vine robot. 
Additionally, to illustrate the ability to create loop closure grasps for objects resting on a surface with no gaps (i.e. the resting surface fully blocks the desired path for closing the loop around the object), a glass vase grasped and lifted from a table, as shown in Fig. 1D and Movie S4.
Three vine robots grow their tips through the interface between the vase and table from opposite directions to balance the horizontal pushing force that each exerts on the object.
The vine robots navigate past the obstruction of the resting surface without a preexisting gap to close the loop and gently hold the vase for lifting.

% \subsubsection*{Versatility enables by tip-everting growth mechanism}

In addition to enabling loop closure grasping, vine robots have benefits for achieving typically difficult grasps due to (1) the inextensibility of its membrane material, and (2) its high length extension ratio. 
The first characteristic is beneficial for bearing tensile loads significantly greater than the payloads of human-safe robots \cite{sherwani_collaborative_2020} (as we later demonstrate in the “Strong and Gentle Grasping and Lifting” section), expanding the versatility of object weights that can be lifted. 
Its inextensibility also enables the transmission of forces over long distances, as initially hypothesized in \cite{hawkes_soft_2017}. 

The high extension ratios of vine robots enable the grasping mechanism to grasp objects that are both (1) substantially far away from its base, and (2) substantially larger than its base. 
This is because, in order to create a stable grasp, the mechanism must be able to sufficiently reach around the object to apply the constraining contacts/forces at the appropriate locations, which may not be feasible for objects significantly larger than the grasping mechanism or far away from the base frame. 

To demonstrate the ability of the vine robot grasping mechanism to grasp and pull objects at distances much further away than its own length, we performed loop closure grasps on boxes 3.0 m away and pulled it to the base, as shown in Fig. 4E and Movie S5, and additional details in Supplementary Materials. 
The distance over which the container was grasped and pulled was 896\% longer than the length of the vine robot base system (0.335 m), with the vine robot growing to a total length of over 6.7 m. 
The combined cross-sectional area of the bounding boxes of the motorized base and tip-fastening winch is 0.180 m$^2$, whereas the cross-sectional area of the larger box in the horizontal plane is 135\% greater at 0.244 m$^2$. 
Having grown over 6.7m long, the vine robot theoretically could have also grasped a larger nearby object with a circular cross-sectional area over 1,985\% greater than that of the base and winch. 

We also demonstrate multi-object grasps (as shown in Fig. 5B and Movie S6) by wrapping around and cinching a pile of pipes by retracting the vine robots' length back into the base while the tip is still fastened. The closed-loop morphology encloses the pipes into a bound space, and tightening the vine robots creates a force closure grasp that secures the pipes together via friction, removing the grasp's dependence on gravity to keep the object in a cradled state. The system then manipulates the grasped objects to a target location, where the vine robots invert to smoothly release the grasp without sliding friction. This approach further highlights the system’s adaptability to handle irregularly shaped or clustered objects while maintaining secure grip and reliable release.

\begin{figure} % Do NOT use \begin{figure*}
	\centering
	\includegraphics[width=\linewidth]{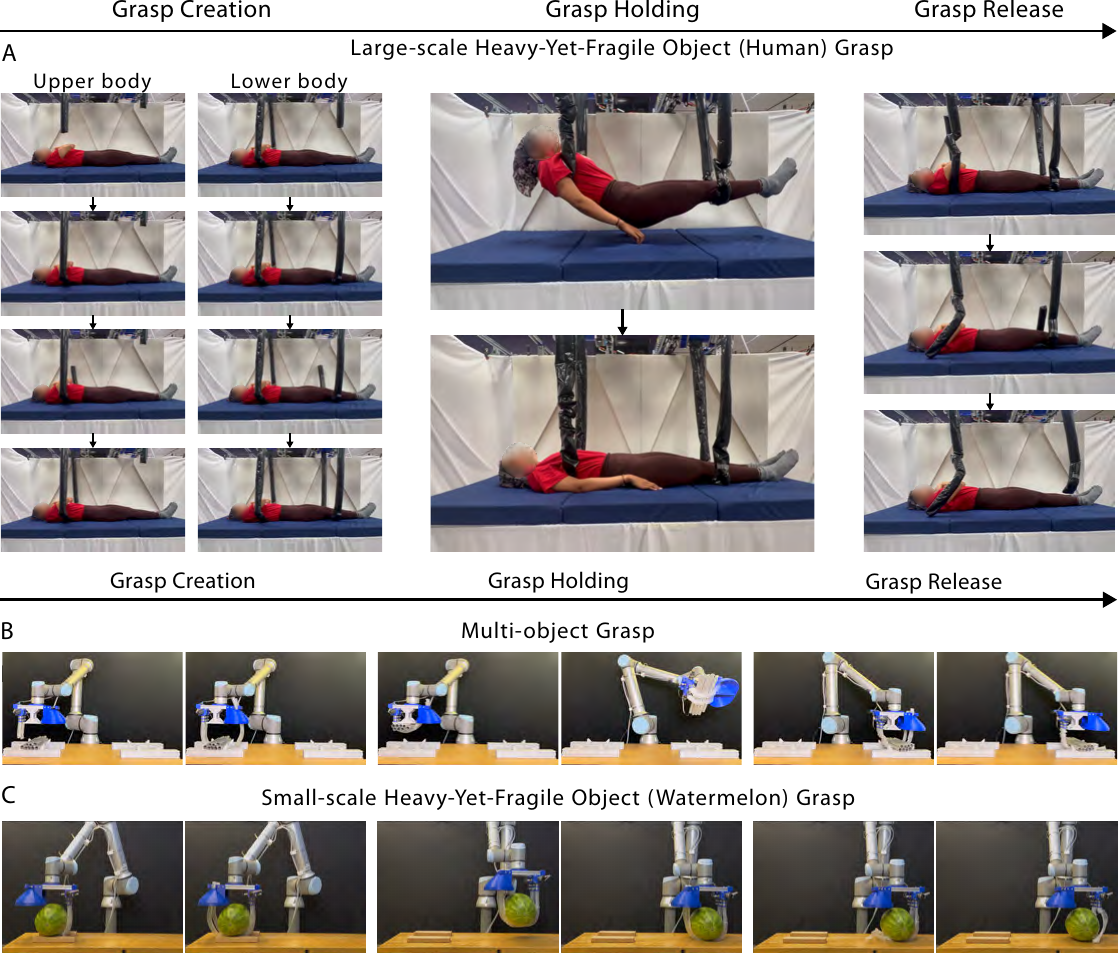} % for an image file named example_figure.*
	% Pick an appropriate width - in print, figures are usually one or two columns wide, which can
	% be approximated by 0.3\textwidth or 0.6\textwidth respectively. Use appropriate label sizes.

	% Captions go below figures
    {\linespread{1.0}
	\caption{
        \textbf{Harnessing and lifting demonstrations.}  \textbf{(A)} Harnessing and lifting a human subject with large-scale grasping implementation system. \textbf{(B)} Simultaneously grasping and transferring multiple objects. \textbf{(C)} Grasping and transferring a heavy yet fragile object (watermelon).}}
	\label{fig:fig5} % give each figure a logical label name
\end{figure}

% \subsubsection*{“In hand” manipulation}

Leveraging the continuous flexibility of the deflated vine robots and the fact that the winches can feed excess length between them, our systems can also manipulate objects “in hand”. 
Assuming sufficient static friction to prevent slipping, the vine robot material can be fed from the base to the tip, or vice versa, while keeping its total length constant such that the object rotates without translating (analogous to a belt driving an idler pulley). 
To demonstrate this, a cylinder was grasped and lifted, and the vine robot was deflated and extended/retracted by the tip and base winches to rotate it while held in mid-air (as shown in Fig. 4C and Movie S7). 

\subsection*{Strong and Gentle Grasping and Lifting}

% \subsubsection*{Human grasping and lifting}

% \begin{figure} % Do NOT use \begin{figure*}
% 	\centering
% 	\includegraphics[width=\linewidth]{Figs/HumanLift-min.pdf} % for an image file named example_figure.*
% 	% Pick an appropriate width - in print, figures are usually one or two columns wide, which can
% 	% be approximated by 0.3\textwidth or 0.6\textwidth respectively. Use appropriate label sizes.

% 	% Captions go below figures
%     {\linespread{1.0}
% 	\caption{\textcolor{red}{Still needs polishing?}
%     \textbf{Harnessing and lifting demonstrations.} \textbf{(A)} Harnessing and lifting a human subject with multiple vine robot grasping systems.}}
%     \label{fig:fig6} % give each figure a logical label name
% \end{figure}

We demonstrate the ability to safely grasp and lift heavy yet fragile objects through loop closure grasping by handling a human with our large-scale system (as shown in Fig. 5A and Movie S8). 
Specifically, two vine robot base/tip device subsystems were fixed above a human (weight: 74.1 kg, height: 170.2 cm) lying on a bed and used to grasp and lift their full weight to illustrate the system’s ability to (1) create a stable enveloping grasp around a body without a pre-existing open path for that configuration, and (2) safely and stably lift a heavy yet fragile object. 

The system was able to close the loop around the subject in a stable configuration (wraps around and supports the body from below) by growing its tip into the interface between the body and the bed, frictionlessly tunneling its way to the other side to place itself under the body, and growing back upward via obstacle-aided navigation \cite{greer_robust_2020} into the tip-fastening winch. 
The system successfully lifted the full weight of the subject approximately 25 cm above the bed without causing any excess harm or discomfort (see Supplemental Materials for participant responses). 
We also demonstrate releasing the grasp by unfastening the vine robot tip and retracting it back from under the body.

To further evaluate the “safety” of using vine robots to create loop closure grasps around humans on a resting surface, we measured the contact pressures experienced by a 79.4 kg life-sized manikin as a single vine robot grew underneath it. 
The manikin was instrumented with a soft capacitive sensor sleeve, and the normal contact pressure distribution was measured over the duration of the vine robot growing under the body (see Methods and Materials for details). 

\begin{figure} % Do NOT use \begin{figure*}
	\centering
	\includegraphics[width=1.0\textwidth]{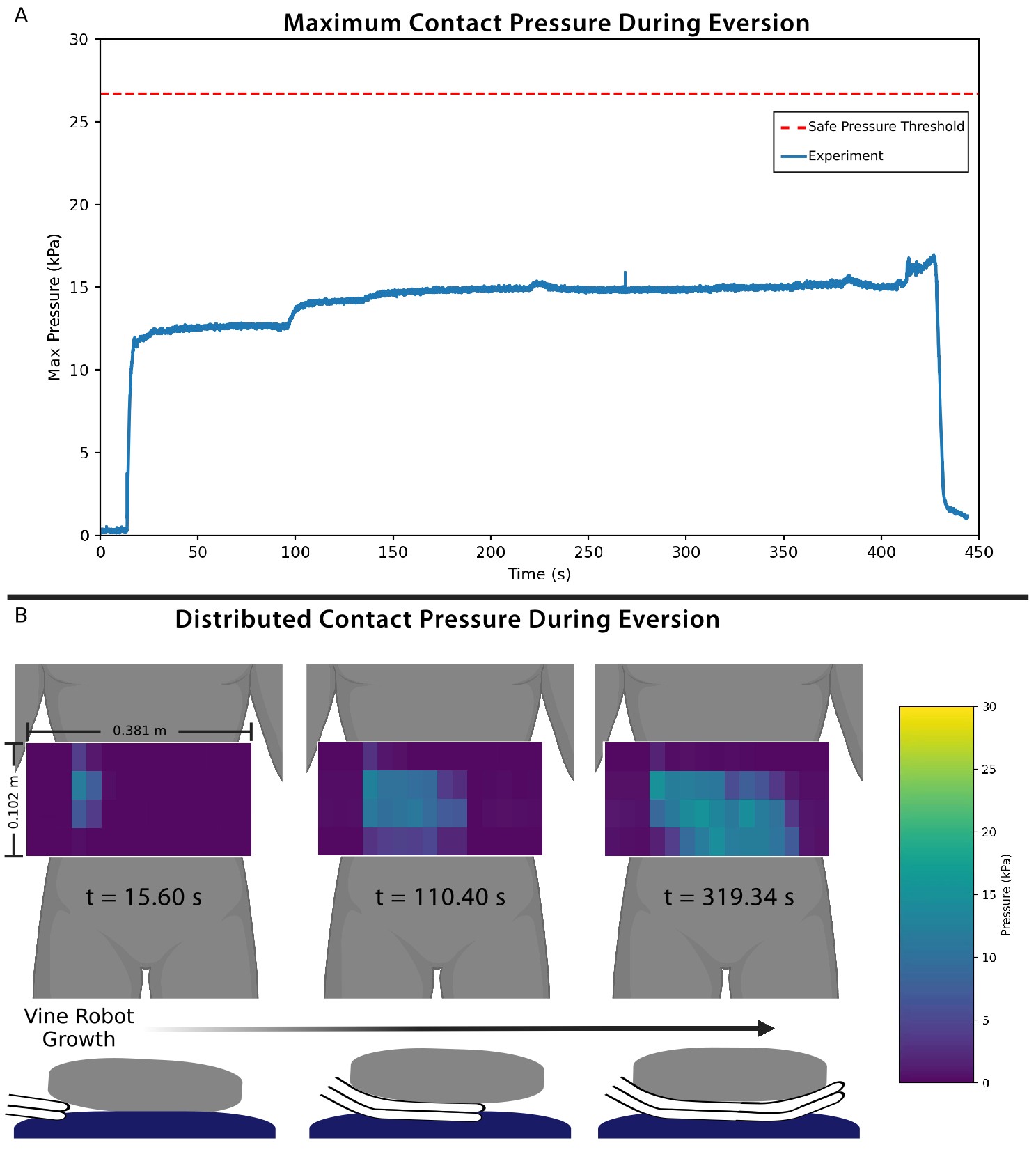} % for an image file named example_figure.*
	% Pick an appropriate width - in print, figures are usually one or two columns wide, which can
	% be approximated by 0.3\textwidth or 0.6\textwidth respectively. Use appropriate label sizes.

	% Captions go below figures
    {\linespread{1.0}
	\caption{\textbf{Pressure distributions from life-sized manikin weight (79.4 kg) lying on bed while vine robot grows under back.} \textbf{(A)} Maximum contact pressure during vine robot growth between the body and the bed. The peak pressure measured over the entire duration was 16.95 kPa. \textbf{(B)} during vine robot growth between the body and the bed at t = 15.60 s, 110.40 s and, 319.34 s. 
    % Pressure distribution when the peak pressure $P_{max}$ = 16.95 kPa is measured (t = 426.90 s). \textbf{(C)} Pressure distribution after the vine robot has grown across the full width of the body and the driving pressure is released (t = 444.25 s).
    }}
	\label{fig:fig7} % give each figure a logical label name
\end{figure}

As shown in Fig. 6 and Movie S9, as expected, the maximum pressure concentrations occur at the contact between the body and the inflated vine robot. 
The peak pressure measured over the entire duration was 16.95 kPa. 
This is significantly lower than the peak pressures experienced by patients during standard transfers in medical patient slings ($>26.7$ kPa) \cite{peterson_pressure_2015}, showing that the interaction with the vine robot during the grasping process is safe for the human body.

% \subsubsection*{Grasping heavy yet fragile objects}

We further demonstrate the ability to safely grasp heavy yet fragile objects with our small-scale system  by grasping and lifting a watermelon (as shown in Fig. 5C and Movie S10) and a 1.0 kg glass vase (as previously described in the Grasp Versatility section, and shown in Fig. 1D and Movie S4). 
For the former, the system was positioned above a 5.9 kg watermelon (29 x 20 x 20 cm), and two vine robots independently grew under and around the watermelon into the tip-fastening mechanisms to close the loops. 
The entire device was then raised to successfully lift the watermelon without damage.

%%%%%%%%%%%%%%%%%%%%%%%%%%%%%%%%%%%%%%%%%%%%%%%%%%%%%%%%%%%%%%%%%%%%%%%%%%%%%%%%%%%%%%%%%%%%%%
\section*{Discussion} 

Loop closure grasping transitions between both open-loop and closed-loop topologies to leverage each of their respective benefits for the different stages of the grasping process. 
This marks a significant departure from traditional design paradigms that only make use of a single topology, in that it relieves the need for the open-loop form to satisfy the functional requirements of maintaining safe and effective grasps, and the need for the closed-loop form to satisfy the functional requirements for creating versatile grasps. 

Decoupling each topology’s involvement from their lesser-suited stage of the grasping process enables their full potential to be utilized to best satisfy the stage they are best suited for. 
This is particularly highlighted by our proof that open-loop grasping mechanisms fundamentally cannot hold objects with “infinite bending compliance” while closed-loop mechanisms can, in that grasping mechanisms designed with permanently open-loop morphologies to enable versatile grasping will always be limited in the compliance it can achieve relative to closed-loop morphologies. 
Conversely, open-loop mechanisms can theoretically navigate around objects through any path that their tip can pass through (as demonstrated in continuum and hyper-articulated robotics, assuming its tip is its widest point,  and growing robotics, as shown in this work and past literature), whereas permanently closed-loop mechanisms are fundamentally hindered by needing to navigate their entire linkage around the object at once due to lacking a free tip. 

By using vine robots with tip-fastening mechanisms, we enable the full benefits of both topologies as well as the ability to transition between them to utilize them in the respective stages of grasping that they are beneficial for. 
Vine robots are particularly beneficial for creating versatile grasps due to their unique ability to navigate through highly constrained environments via eversion-based growth, including through spaces that are smaller than its tip (as shown in literature \cite{hawkes_soft_2017, blumenschein_design_2020, coad_vine_2020}, our cluttered environment and human lifting demonstrations) and their highly articulable continuum structure. 
Their high extension ratio coupled with their compact storability also enables scalable grasping of larger objects, with significant increases in available vine robot length only requiring small increases in the size of the base. 
Their light weight and soft structure allows for safe navigation around fragile objects as well. 
Due to the flexible and inextensible construction of vine robot membranes, their morphology changes from a tube to a sheet when deflated. 
Like with slings, belts, straps, and hammocks used in industry, this morphology lends well for achieving high tensile strength and virtually zero flexural rigidity, enabling it to fully leverage the closed-loop topology when its tip is fastened for safely holding high loads with “infinite compliance”. 

Our method and system design principles can theoretically be realized using any type of robotic open-loop mechanism that can fasten its tip to its base \cite{jiang_vacuum_2021, 
% morimoto_design_2016, 
wang_design_2021, becker_active_2022, kaneko_active_1998}. 
Different mechanisms can present different advantages and disadvantages, and thus this work presents loop closure grasping as a general method that can be implemented with any type of mechanism that satisfies our proposed architecture and principles. 
While vine robots are particularly beneficial for realizing loop closure grasping, they also present some challenges including easily disturbed growth trajectory away from the desired path toward the tip-fastening mechanism and difficult to instrument with sensors for closed-loop position control. 
Investigations of loop closure grasping system designs utilizing different mechanisms can further elucidate best practices for implementation in different applications with different goals and requirements. 

We demonstrate how our loop closure grasping paradigm enables versatile grasping of objects that are heavy yet fragile, topologically challenging, large in scale, and in constrained environments, all of which have traditionally been challenging for robots. 
In addition to advancing performance, these benefits are also conceptually significant for soft robotic grasping, in that the “infinite compliance” maximizes one of the core benefits of soft robotics, and the high pull-out force overcomes one of its most significant traditional limitations (compared to rigid grasping mechanisms). 
Further, we envision that this paradigm will broaden the practical applicability of robotic grasping to a wider range of potential high-impact applications. 
Robotics could be made more applicable to many industries that require handling of humans, such as eldercare, care for people with disabilities, patient care in hospitals, emergency medical response, and search and rescue. 
While our current demonstration of lifting a human with only two vine robots may not provide the most comfortable experience, future work can investigate the use of more, wider, and/or branching vine robots to distribute the load over larger contact areas. 
Other applications involving heavy yet fragile objects include handling livestock, agricultural harvesting, heavy manufacturing, large cargo handling, construction, infrastructure repair, and ship/aircraft wreckage recovery. 
Grasping versatility is important for these applications as well, as the robot must adapt to the various requirements for each presented task. 
In aerospace applications, long-extending loop closure grasping mechanisms could be used to collect in-orbit space debris, transfer payloads between spacecraft, or re-tether detached astronauts. 
In wearable applications, loop closure mechanisms could be used to automatically fasten and secure clothing and devices to the body. 
In fast-paced warehouse sorting, the passive strength, stability, and compliance of loop closure grasps could enable more dynamic manipulation by mitigating the effects of potentially destabilizing or damaging high inertial accelerations. 

To transition from the open-loop to closed-loop topologies, the tip of the open-loop mechanism must be able to navigate its tip to the base to be fastened. 
While we successfully demonstrate this using preforming methods to program the path through which the vine robots grow, a limitation of our study is that we did not utilize actuated steering or bending mechanisms to actively articulate its tip into the fastening mechanism. 
Adding active articulation (which has been studied extensively for robots with open-loop morphologies \cite{blumenschein_design_2020} could dramatically improve the robustness and adaptability of loop closure grasping systems to the levels of dexterity enabled by current continuum and hyper- articulated robots. 
Robust implementations of active articulation of vine robots in 3D at larger scales remains an open research question that stands to greatly improve their use in loop closure grasping.
Additionally, the safety of interaction between our systems and the objects (e.g. a human) while bearing their weight could be further investigated. 
In-depth theoretical analyses of the pressure distributions and contact mechanics of loop closure grasping systems could be further elucidated and enable the optimization of future grasping mechanisms using this paradigm.

%%%%%%%%%%%%%%%%%%%%%%%%%%%%%%%%%%%%%%%%%%%%%%%%%%%%%%%%%%%%%%%%%%%%%%%%%%%%%%%%%%%%%%%%%%%%%%
\section*{Materials and Methods} 

\subsection*{Construction of loop closure grasping system components}

The vine robots used in our large-scale systems are constructed by heat-sealing the thermoplastic polyurethane (TPU)-coated sides of 70-denier nylon ripstop fabric (Seattle Fabrics Inc.) in a tubular shape to provide a high tensile strength, airtight, inexpensive and lightweight construction with negligible flexural rigidity. 
For the small-scale systems, we construct the vine robots using low-density polyethylene (ULINE), which has a lower but still sufficient tensile strength of 10 MPa \cite{szlachetka_low-density_2021}. 
We utilize preforming to navigate the vine growth direction, as well as obstacle-aided passive steering for the human grasping demonstration (both are standard methods established in vine robot literature \cite{blumenschein_design_2020}). 

The base consists of a motorized winch inside of a pressurized chamber with an air-tight collar to fasten the base of the vine robot (Fig.3).
For the large-scale systems, the chamber is made from aluminum to provide a strong grounding frame on which the winch can exert high pulling forces. 
The collar and winch are 3D printed out of carbon-fiber infused nylon with fiberglass continuous fiber reinforcements.
The winch has a steel shaft core and is driven by a brushless DC motor (Rev Robotics LLC) with a 100:1 ratio gearbox. 
For the small-scale systems, the chamber and collar are 3D printed out of PLA as one piece. 
The winch has a steel shaft core and is driven by a brushed DC motor with a 455:1 ratio gearbox (Robotzone LLC). 

Our implementation of the tip-fastening mechanism for the large-scale system is a winch with an embedded clamp with smoothly corrugated jaws to fasten and wind up load-bearing vine robots (Fig. S5). 
The tip-fastening winch is constructed primarily from 3D printed PLA, aluminum plates, a steel hex shaft, and two linear actuators (Eco-Worthy), with partially-threaded bolts used to reinforce the 3D printed parts. 
For the small-scale system, both clamps and tip-fastening winches are used for different demonstrations (Fig.3C). 
They are both constructed primarily from 3D printed PLA and a single linear actuator (UYGALAXY). 
See Supplementary Materials for additional material specifications and construction details for all components.

\subsection*{Human grasping and lifting demonstration participant}

A healthy female adult (age, 25 years; weight, 74.1 kg; height, 170 cm) participated in the experiment (Fig.5A). 
The participant was informed about the experiment and provided informed consent before participation. 
The vine robots were pressurized to 24.1 kPa and turned ~$90^{\circ}$ before navigating under the participant, and was pressurized to 13.8 kPa and turned ~$45^{\circ}$ after emerging from the opposite side. 
They grew into the tip-fastening winches via obstacle-aided navigation \cite{greer_robust_2020} using surface features in the adjacent wall. 
The vine robots were then fully depressurized, wound up by both ends to lift and hold the participant 25 cm above the mattress, and unwound to gently lower her back down. 
Prior to the lift, the participant was instructed to keep their body slack except for raising their head to avoid injury as they were lowered. 
The participant's upper body is lifted first to ensure an upright position. 
Then, their lower body is lifted.
The participant rated her comfort throughout the experiment using Likert scales. 
All experimental procedures were approved by the respective Institutional Review Boards of the Massachusetts Institute of Technology and Stanford University.

\subsection*{Contact pressure  during growth under human experiment}

For the experiment illustrated in Fig.6, a 79.4 kg life-sized manikin (Simulaids I.A.F.F. Rescue Randy) was used. Its weight is distributed to match the average US adult weight distribution. To measure contact pressures, it was outfitted with a custom soft capacitive sensor array created by Pressure Profile Systems (Los Angeles, CA, US). See Supplementary Materials for further details.

% If your text is very short you might need to uncomment the following line to avoid
% layout problems with the figures and tables.
%\newpage

%%%%%%%%%%%%%%%% REFERENCES %%%%%%%%%%%%%%%

\clearpage % Clear all remaining figures and tables then start a new page

% The list of references goes after the main text and before the acknowledgements
% When preparing an initial submission, we recommend you use BibTeX, like this:
%
\bibliography{SciRob_2024_fullText.bib} % for a file named science_template.bib
\bibliographystyle{sciencemag}

% After the paper has completed peer review and been revised ready for acceptance,
% you should comment out the lines above and copy-paste the contents of your .bbl
% file here instead. This will help ensure that our conversion software works correctly.
% Remember to re-run BibTeX first - check the timestamp!
%
% Example of the first three entries copy-pasted from science_template.bbl:
%
%\begin{thebibliography}{1}
%
%\bibitem{example}
%A.~N. {Author}, An example reference. \emph{Journal of Improbable Research}
%  \textbf{1}, 67 (2020).
%
%\bibitem{example2}
%F.~M. {Surname}, S.~{Author}, A second example. \emph{Interesting Research
%  Letters} \textbf{32}, 897 (2019).
%
%\bibitem{example_preprint}
%P.~{One}, P.~{Two}, P.~{Three}, {An unpublished preprint}. \emph{preprint}
%  (2021), arXiv:2101.12345.
%
%\end{thebibliography}

%%%%%%%%%%%%%%%% ACKNOWLEDGEMENTS %%%%%%%%%%%%%%%

\section*{Acknowledgments}
We thank Kemi Chung and Nicholas Cerone for their help with the construction of the high-strength lifting system, and Delaney Benevides for her help with performing the multi-pipe and vase grasping demonstrations. 

\paragraph*{Funding:}
National Science Foundation grants \#2344314 (KB, SK, CdP, AMO, HHA) and \#2024247 (AMO). Ford Foundation Predoctoral Fellowship (OGO). 

\paragraph*{Author contributions:}
Conceptualization: (KB, OGO). Data Curation: (KB, OGO, SK, CdP, CMH). Formal Analysis: (KB, OGO, CdP, CMH). Funding Acquisition: (HHA, AMO). Investigation: (KB, OGO, SK, CdP, CMH). Methodology: (KB, OGO, HHA, AMO). Project Administration: (KB, OGO, HHA, AMO). Resources: (HHA, AMO). Software: (KB, OGO, SK). Validation: (KB, OGO). Visualization: (KB, OGO, SK, CdP, CMH). Writing – Original Draft: (KB, OGO, CdP). Writing – Review \& Editing: (KB, OGO, HHA, AMO).

\paragraph*{Competing interests:}
Authors declare that they have no competing interests.

\paragraph*{Data and materials availability:}
All data are available in the main text or the supplementary materials.

% %%%%%%%%%%%%%%%% SUPPLEMENT LIST %%%%%%%%%%%%%%%

% % List the contents of your Supplementary Materials, including the numbers of any
% % supplementary figures, tables, external data files etc. and any references that are
% % cited only in the supplement. In this example, refs. 7-8 are cited only in the supplement.
% % Fill out your numbers accordingly and delete any lines that aren't applicable.
% \subsection*{Supplementary materials}
% Supplementary Text\\
% Figs. S1 to S4\\
% References \cite{rodriguez_caging_2012, hanafusa_stable_1977, brown_universal_2010}\\ % automatically fills out the last reference number
% % (filling out the other numbers automatically is possible but fiddly and liable to break)
% Movie S1 - S8\\
% Data S1 - S3

%%%%%%%%%%%%%%%% END OF MAIN TEXT %%%%%%%%%%%%%%%

\newpage

%%%%%%%%%%%%%%%% START OF SUPPLEMENT %%%%%%%%%%%%%%%

% Figures, tables, equations and pages in the supplement are numbered S1, S2 etc.
\renewcommand{\thefigure}{S\arabic{figure}}
\renewcommand{\thetable}{S\arabic{table}}
\renewcommand{\theequation}{S\arabic{equation}}
\renewcommand{\thepage}{S\arabic{page}}
\setcounter{figure}{0}
\setcounter{table}{0}
\setcounter{equation}{0}
\setcounter{page}{1} % not 0 as \newpage already started a supplementary page
% References continue the numbering from the main text.

\begingroup
\singlespacing

%%%%%%%%%%%%%%%% SUPPLEMENT TITLE PAGE %%%%%%%%%%%%%%%

\begin{center}
\section*{Supplementary Materials for\\ \scititle}

% Author list for the supplement
% Indicate the corresponding authors, but do NOT include institutions here
% It would be nice if the template auto-generated this, but doing so is complicated...
Kentaro~Barhydt$^{1\dagger\ast}$,
O.~Godson~Osele$^{2\dagger}$,
Sreela~Kodali$^{2}$,
Cosima~du~Pasquier$^{2}$,
Chase~M.~Hartquist$^{1}$, \and
H.~Harry~Asada$^{1}$,
Allison~M.~Okamura$^{2}$ \\
% Additional lines of authors should be inserted using the \and command (not \\)
% Institution list, in a slightly smaller font
\small$^{1}$Department of Mechanical Engineering, Massachusetts Institute of Technology, Cambridge, MA 02139, USA. \\
\small$^{2}$Department of Mechanical Engineering, Stanford University, Stanford, CA 94305, USA. \\
% Identify at least one corresponding author, with contact email address
\small$^\ast$Corresponding author. Email: kbarhydt@mit.edu\\
% Joint contributions can be indicated like this
\small$^\dagger$These authors contributed equally to this work and are co-first authors.
\end{center}

% Fill out the numbers for each type of supplementary material,
% and delete any lines that aren't applicable.
% These are just example numbers that don't match the rest of this template.
\subsubsection*{This PDF file includes:}
Supplementary Discussions S1 to S9\\
Figs. S1 to S5\\
Labels for Movies S1 to S10

\subsubsection*{Other Supplementary Materials for this manuscript:}
Movie S1 - S10\\
Data S1 - S3

\newpage

%%%%%%%%%%%%%%%% MATERIALS AND METHODS %%%%%%%%%%%%%%%

\subsection*{Supplementary Discussion S1. Additional context for problem definitions section}

Here we present a formalization of the task of harnessing (i.e. grasping in loop closure, grasping in weak form closure or in a grasping cage \cite{rodriguez_caging_2012}) and pulling heavy yet fragile objects. 

The key factors for success are the security and gentleness of the grasp. 

\begin{itemize}
    \item \textbf{Grasping}: We consider the object to be grasped if it is mechanically stable inside of the grasp such that some level of work is required to remove it from the stable region, and the security is quantified as the amount of work required to destabilize it \cite{hanafusa_stable_1977}. The grasp must be performed independently, in that no external physical effort from either the object or an outside agent is required. 
    \item \textbf{Pulling}: We consider the object to be pulled if the subsequent manipulation forces displace it toward the base of the manipulator over some distance. 
    \item \textbf{Gentleness}: We quantify the “gentleness” of the interaction as the ratio between the maximum pressures exerted onto the object and the maximum allowable contact pressures the object can experience without harm/damage.
    \item \textbf{Heavy Yet Fragile}: We define a heavy yet fragile object as one with a significantly high weight relative to its allowable interaction forces, and quantify this characteristic as the ratio of its weight to its maximum allowable contact pressures. 
\end{itemize}

To address the versatility of applications in which a mechanism can achieve this task, we also define grasping versatility in terms of three factors: 

\begin{itemize}
    \item (1) the variety and scale of objects it can grasp,
    \item (2) the variety of grasping configurations it can deploy (i.e. configuration space of the grasping mechanism), and
    \item (3) the variety of environments within which it can grasp these objects.
\end{itemize}

The second and third factors are included to account for the fact that a mechanism’s ability to successfully grasp an object is also dependent on the accessibility of the object in the given environment and the desired grasping configurations for the given task.

\newpage
\subsection*{Supplementary Discussion S2. Additional context for closed-loop vs. open-loop morphologies}

We present a general framework for representing grasping mechanisms in the “Closed-Loop vs. Open-Loop Morphologies” section. Additionally, For grasping mechanisms with branching kinematic chains, we consider each chain, starting at one of the tips and ending at either the base or a point along another chain, to be its own grasping mechanism. Thus the number of mechanisms equals the number of tips. If the full geometric paths that the grasping mechanism creates relative to the object cannot be represented with a single one-dimensional serial chain (e.g. textiles), then it can instead be abstracted as a two-dimensional manifold (a surface), with its base and tip taking the same definitions as the grounded points and distal points of contact, respectively. Again, because we are interested only in the paths that the mechanism creates relative to the object to bridge the base and its points of contact, paths created by mechanisms with kinematics that could be represented with a three-dimensional geometry (e.g. particle jamming gripper \cite{brown_universal_2010}) can be simplified as just a one- or two- dimensional manifold for our definition.

We also define the closed-loop morphology as meeting the following two criteria: (1) the tip is grounded to the same system as the base, and (2) the object is inside of the loop created by the mechanism. For the first criterion, if the tip (the most distal contact point along the mechanism) is defined by its fixture to the base instead of contact with the object, then the path through the kinematic chain of the grasping mechanism (including the base) is now closed. If the tip is grounded to a different frame than the base, it still satisfies criteria one as long as the positions of both grounds are fixed relative to each other or manipulated relative to each other by the same system (e.g. both fixed to the world frame, both fixed to the same manipulator, different manipulators on same robot, different robots in the same multi-robot control system). The ground segment that closes the kinematic chain from the tip to the base is defined by the shortest path between them. We include the second criterion in our definition because the grasping morphology is concerned with how the mechanism interacts with the object, not just with the mechanism itself. Many grasping mechanisms include closed-loop kinematic chains within them (e.g. parallel linkage gripper), but the object is not inside of those closed loops, and thus their tips are not grounded and are still considered to have an open-loop grasping morphology. Closed-loop mechanisms surround the object in at least one plane, which creates a fully caged grasp in the two-dimensional space of that plane. Similar to how multiple open-loop grasping mechanisms can be used in tandem to create a fully caged grasp in three-dimensional space \cite{rodriguez_caging_2012}, multiple closed-loop mechanisms can be used to accomplish this as well. To the best of the authors’ knowledge, the only examples of robotic mechanisms utilizing closed-loop morphologies presented in literature were presented in \cite{barhydt_high-strength_2023, kang_grasping_2023}.

\newpage
\subsection*{Supplementary Discussion S3. Closed-loop grasping morphologies enable grasps with infinite bending compliance (proof)}
\label{sec:Supp_infiniteComplianceProof}

Here we present a logical proof that open-loop grasping mechanisms with zero flexural rigidity cannot stably pull on objects, whereas closed-loop mechanisms with zero flexural rigidity can. 
Open-loop mechanisms constrain the object through geometric and/or force constraints \cite{tai_state_2016, 
% bicchi_robotic_2000, 
prattichizzo_grasping_2016}. 
We consider the two respective types of grasping: enveloping (caging, form closure) and pinching (force closure), as shown in Fig. S1A and Fig. S1E.

In the enveloping case, the grasping mechanism pulls on the object by applying a force $F$ behind it. 
To reach behind it, the mechanism must extend past the side of the object. 
Thus, a moment arm $d$ will always exist between the base of the mechanism and the point of force application, creating an internal bending moment $M$. 
This applies for all mechanisms, open- and closed-loop. 
For open-loop morphologies, if the mechanism has no flexural rigidity, it cannot resist the moment $M$, and thus will bend away from the object until it is no longer behind it. 
Some level of flexural rigidity, whether structural or induced through actuation, is required to resist $M$. 
Similar logic can be used to evaluate the pinching case (Fig.S1E-F), in which open-loop mechanisms generate friction forces $F_f$ in the pulling direction by applying normal forces $N$. 
The object must be pulled over some distance, so a moment arm $d$ will always exist. 
Without any flexural rigidity, the open-loop mechanism cannot resist the resulting moments $M$ and will bend away from the object until it is no longer in contact. 
Therefore, open-loop mechanisms with zero flexural rigidity cannot stably hold and pull grasped objects. 

\begin{figure} % Do NOT use \begin{figure*}
	\centering
	\includegraphics[width=1.0\textwidth]{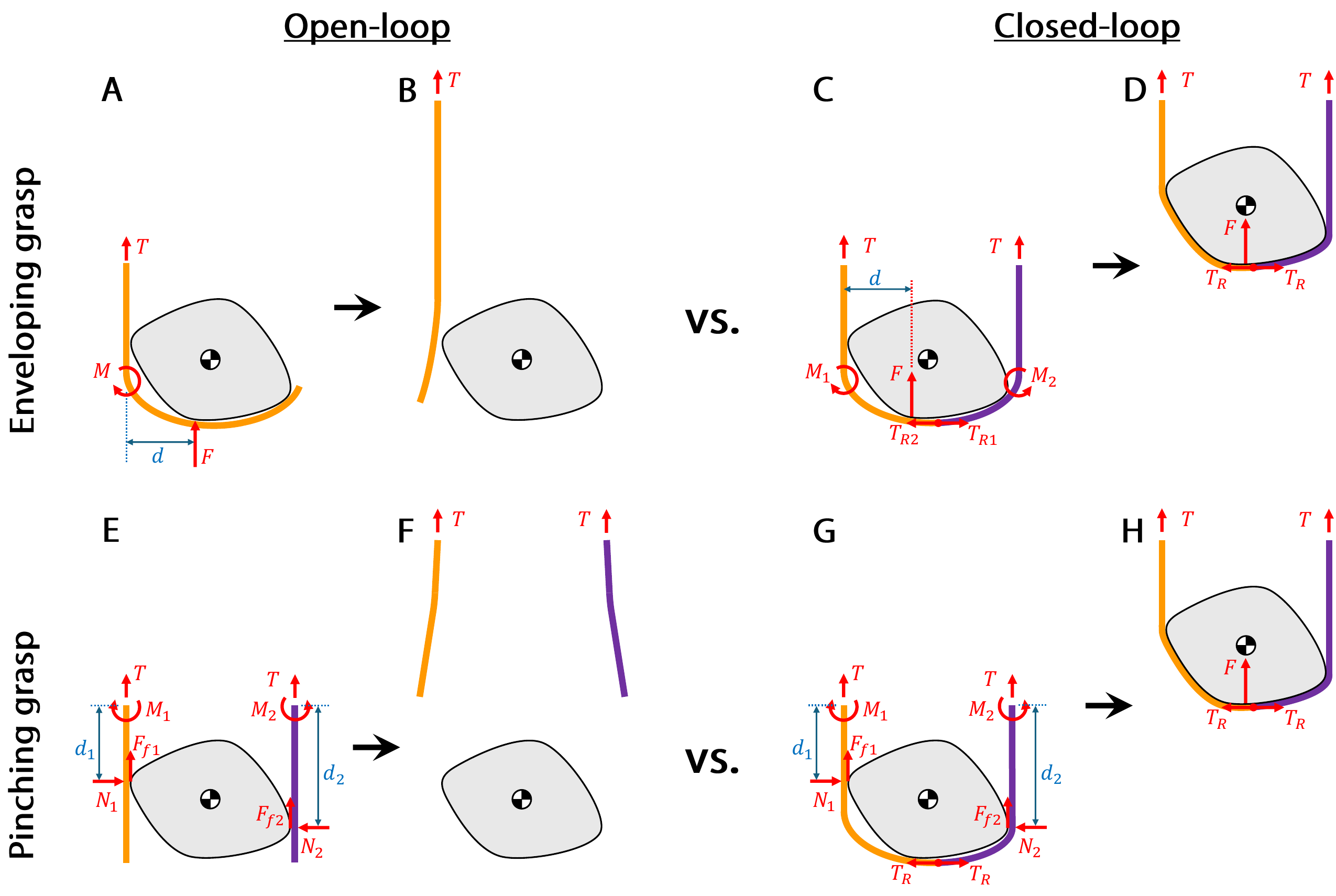} 
     {\linespread{1.0}
	\caption{\textbf{Closed-loop vs. open-loop grasping morphologies.} \textbf{(A-H)} Force diagrams of open-loop grasping mechanisms before and after pulling on objects for enveloping grasping and pinching grasping, and their associated closed-loop mechanism comparisons. $T$, $M$, $T_R$, $F$, $F_f$, $N$, and $d$ denote pulling tension, internal bending moment, internal reaction tension, pushing force, frictional force, normal pinching force, and moment arm, respectively. }}
	\label{fig:infComplianceProof1} % give each figure a logical label name
\end{figure}	

For comparison in both cases, we evaluate closed-loop mechanisms as two pseudo-open loops on either side of the object with their tips fastened together (Fig. 2C-D, Fig. S1G-H). 
The segmentation point between them can be chosen arbitrarily. 
If these pseudo-open loops apply pushing forces $F$ or frictional forces $F_f$ to pull the object, bending moments $M$ will still be generated. 
However, if a pseudo-open loop mechanism cannot independently resist M and begins to bend away from the object, the other mechanism is pulled further around the object. 
Therefore, even if the mechanisms have no flexural rigidity and deform due to contact forces, they will always surround the object in a stable caged grasp.

\subsubsection*{Additional supporting force analysis}

\begin{figure} % Do not use \begin{figure*}
	\centering
	\includegraphics[width=0.4\textwidth]{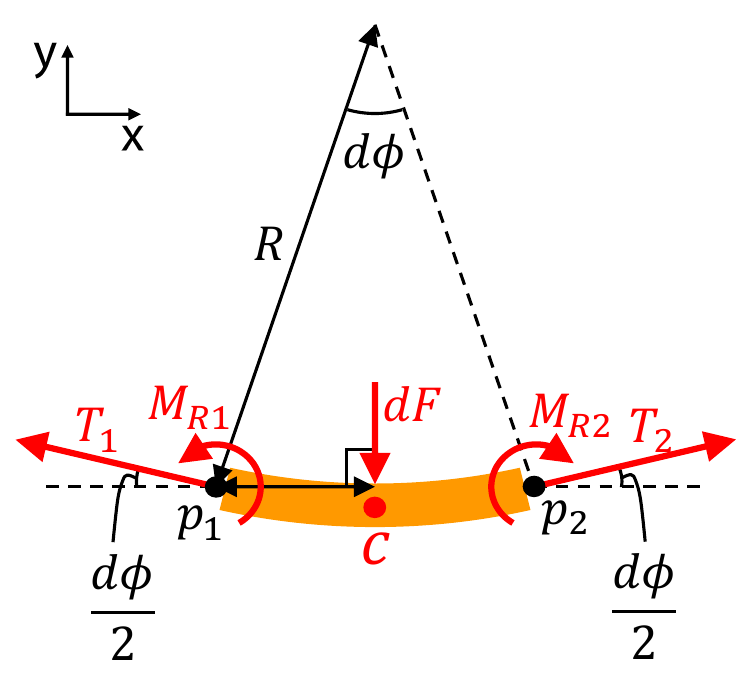} % for an image file named example_figure.*
	% Pick an appriopriate width for the size of the image

	% Captions go below figures
    {\linespread{1.0}
	\caption{\textbf{Free body diagram for a differential segment of a grasping mechanism.} Differential segment associated with contact point c.}}
	\label{fig:infComplianceProof2} % give each figure a logical label name
\end{figure}

Here, we provide further support of this proof through a supplementary analysis of a differential segment of a grasping mechanism. 
As the closed-loop mechanism is pulled, it fully conforms to the object and becomes taught. 
Here, we show that the reaction forces from the base pulling both ends of the loop in tension can balance the internal moments, enabling stable pulling. 
Fig. S2 shows a free body diagram of a differential segment of a grasping mechanism at one of its contact points with the grasped object. dF denotes the differential normal force from the object contact, c denotes the effective contact point at which dF is applied, R denotes the radius of curvature, $p_1$ and $p_2$ denote the end points at the left and right of the differential segment, respectively, $d\phi$ subtends arc $p_1$ $p_2$, $T_1$ and $T_2$ denote the tensile forces applied onto the differential segment from the adjacent segments at points $p_1$ and $p_2$, respectively, and $M_{R1}$ and $M_{R2}$ denote the bending reaction moments in the differential segment due to its flexural rigidity EI, respectively. The subscripts 1 and 2 denote the side closer to the base and tip, respectively. $M_{R1}$ and $M_{R2}$ are generated in reaction to the change in local curvature, i.e. bending with respect to the adjacent segments. As the force of the object onto the mechanism pushes it to bend back, the curvature of the mechanism decreases, causing the reaction moment $M_R$ to increase according to the bending moment equation: $M_R=(\kappa-\kappa_0)EI$. 

The balances of the force and moment about point $p_1$ are: 
\begin{equation}
	\sum F_x = (T_2 - T_1) \cos{\frac{d\phi}{2}}
\end{equation}
\begin{equation}
	\sum F_y = (T_1 + T_2) \sin{\frac{d\phi}{2}}-dF
\end{equation}
\begin{equation}
	\sum M_{p1} = M_{R1} - M_{R2} + (T_2 \sin{\frac{d\phi}{2}}) (2 R \sin{\frac{d\phi}{2}}) - dF (R \sin{\frac{d\phi}{2}})
\end{equation}
For the grasp to be stable, the forces and moments of all differential segments of the grasping mechanism must be balanced when at steady-state. We can therefore ignore inertial effects. 

For open-loop grasping mechanisms, $T_2=0$ and because the tip of the mechanism is free, so there is no tension pulling the segment toward the tip. Therefore, the moment balance becomes: 
\begin{equation}
	\sum M_{p1} = M_{R1} - M_{R2} - dF (R \sin{\frac{d\phi}{2}})
\end{equation}
If the mechanism has zero flexural rigidity ($EI=0$), then $M_{R1}=M_{R2}=0$, yielding:
\begin{equation}
	\sum M_{p1} = - dF (R \sin{\frac{d\phi}{2}})
\end{equation}
The moment is not balanced (unstable) as long as some contact force is applied. Therefore, open-loop mechanisms with zero flexural rigidity cannot stably pull on objects. 

However, in closed-loop mechanisms, the tip is grounded, and thus can supply a reaction tensile force $T_2$ to the entire mechanism. Thus, for a differential segment, the force and moment balances take the form of equations S, S3, and S4. If the mechanism has zero flexural rigidity and $M_{R1}=M_{R2}=0$, the total moment can still be balanced ($\sum M_{p1} = 0$).

\clearpage
\subsection*{Supplementary Discussion S4. Closed-loop vs. open-loop mechanism pressure distribution finite element analysis}

The finite element models used for this work were built in Abaqus CAE 2021. The models are 2D Deformable bodies with a Solid, Homogeneous section. The closed-loop with negligible flexural rigidity was modeled using an Elastic material model with directional rigidity, where the stiffness along the loop (E1) was set to 125 MPa, and the flexural rigidity (E2) was set to 0.1 MPa. The closed-loop mechanism with flexural rigidity and open-loop mechanism with flexural rigidity both used isotropic, non-directional material models with 125 MPa stiffness. The stiffness of the sphere was set to that of steel for all three cases (E = 200 GPa). The mass density for the loops was set to $5\times10^{-9}$ tonne/mm$^3$ for the loops and $7.9\times10^{-9}$ tonne/mm$^3$ for the spheres. We used 2D tetrahedral meshes for both bodies, with a global size of 20 mm. 

The simulation was run in two separate steps, both Static General. In the first step, we used thermal expansion to ensure that the simulation would converge while the sphere was in full contact of the loops. For all three simulations, we used an expansion coefficient of $5\times10^{-4}$ on the steel spheres and applied a Predefined Temperature Field with a magnitude of 8.2 which we ramped throughout the step. We used General Contact on both bodies without friction and using “Hard” Contact for normal contact. In the second step, we applied gravity on both bodies using a Smooth Step. We applied Encastre boundary conditions on the fixed ends of the loops.     

In post processing, we extracted CPRESS, the magnitude of the net contact normal force, from the last frame of each simulation. 

\begin{figure} % Do not use \begin{figure*}
	\centering
	\includegraphics[width=1.0\textwidth]{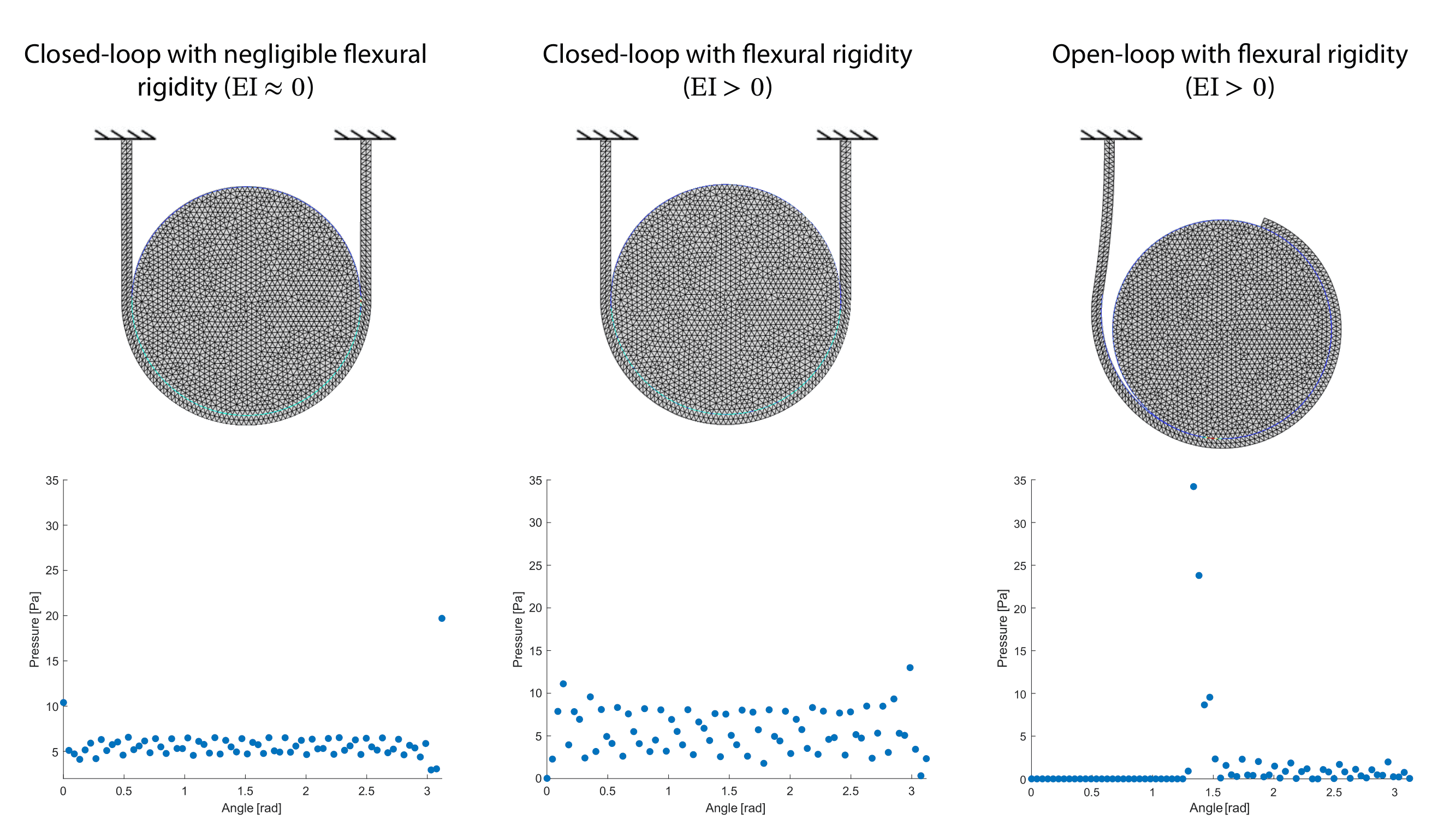} % for an image file named example_figure.*
	% Pick an appriopriate width for the size of the image

	% Captions go below figures
    {\linespread{1.0}
	\caption{\textbf{Closed-loop vs. open-loop mechanism pressure
distribution finite element analysis.} Comparison of contact pressure profiles for an object grasped by a closed-loop mechanism with elastic modulus $E\approx0$ (left), a closed-loop mechanism with E=125 MPa (center), and an open-loop with E= 125 MPa (right).}}
	\label{fig:FEA_OpenLoopPressureDist} % give each figure a logical label name
\end{figure}

% \newpage
\clearpage
\subsection*{Supplementary Discussion S5. Finite element analysis of effects of varying flexural rigidity of a Closed-Loop mechanism on its pressure distribution on idealized object}

Closed-loop mechanisms holding circular objects were simulated in using finite element methods to observe the relationship between the contact pressure distribution and the flexural rigidity of the mechanism. All simulations were developed and performed in Abaqus. The mechanisms were modeled as a two-dimensional deformable planar shell, with a neutral radius of 0.51 m, an initial radius of 0.5 m, a constant second moment of inertia, and a Poisson’s ratio of 0.2. The Young’s modulus was parametrically swept over the range of mechanisms simulated, from 10 kPa to 10 GPa. Their densities were set to scale linearly with Young’s modulus (based on typical scaling behavior from experiments, i.e., as shown on a standard Ashby plot) with values ranging from $101$ kg/m$^{3}$ to $106$ kg/m$^{3}$. The object was modeled as a 2D rigid body discrete wire object with a radius of 0.45 m. The contacts were defined as interactions between the top surface of the mechanism and the perimeter of the object. All contact interactions are set to be frictionless. Contact interactions were tracked – including pressures, forces, and areas (length in 2D) – node-wise across the top/inner surface of the mechanism as a field output for post-processing. A fixed-fixed semicircle beam boundary condition was applied. The downward force was also applied uniformly to the boundary of the object, ramped from zero to the final weight. For the solver, static steps were implemented throughout the closed loop simulation procedure. 

\begin{figure} % Do not use \begin{figure*}
	\centering
	\includegraphics[width=1.0\textwidth]{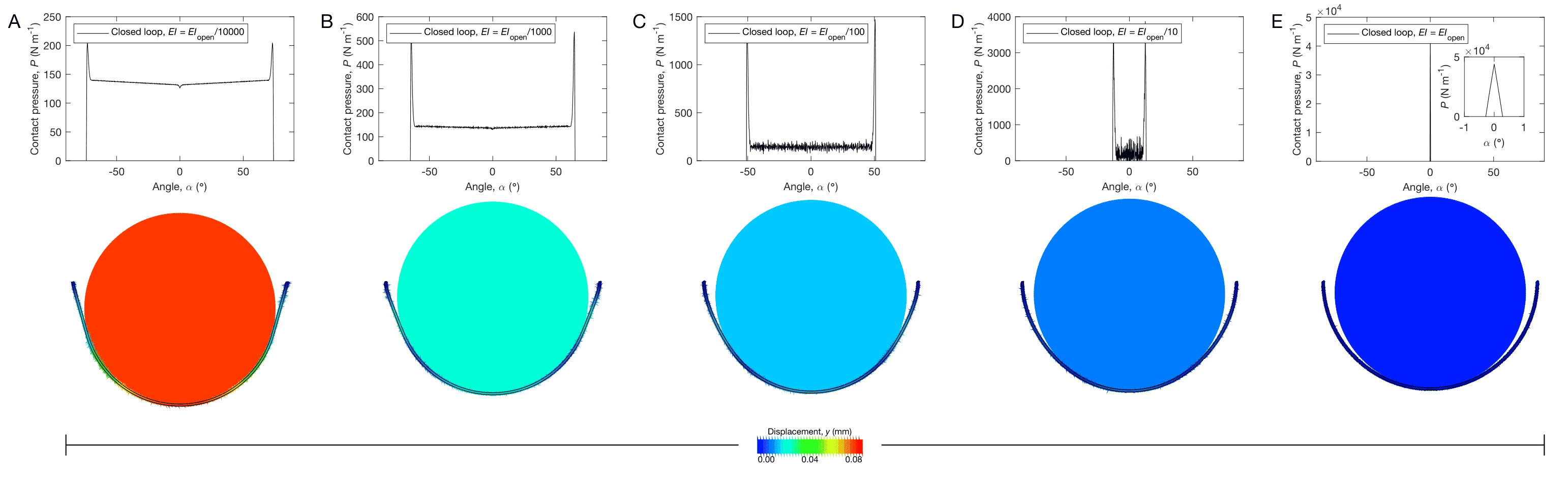} % for an image file named example_figure.*
	% Pick an appriopriate width for the size of the image

	% Captions go below figures
    {\linespread{1.0}
	\caption{\textbf{Pressure distributions between a circular object and five different closed-loop mechanisms with different flexural rigidities.} As flexural rigidity increases, the contact pressure concentrations at the touch-down and lift-off points become greater, increasing the maximum pressure on the object. }}
	\label{fig:FEA_closedLoopPressureDist} % give each figure a logical label name
\end{figure}

\clearpage
\subsection*{Supplementary Discussion S6. Loop closure grasping mechanism implementation details}

The loop closure grasping system concept we describe is comprised of three main components: (1) the base, (2) the grasping mechanism(kinematic chain that transitions from open-loop to closed-loop topology), and (3) the tip-fastening mechanism grounded to the same body and/or system as the base (Fig. 1C). The grasping mechanism navigates its tip around the object into the tip-fastening mechanism to create a loop closure grasp. Although this framework can be realized through many different implementations, we describe the one used for the system shown in Fig. 3, which uses vine robots as the grasping mechanism body, as well as winches in both the base and tip-fastening mechanism to retract the length of the closed loop to apply lifting forces. 

For the demonstrations using the large scale loop closure grasping system, we selected a commercially available thermoplastic polyurethane (TPU) -coated 70-denier nylon ripstop fabric (Seattle Fabrics Inc., WA, USA) for its high tensile strength (1.31x104 N/mm), light weight (170 g/m$^2$), negligible flexural rigidity, and the ability to heat-seal its TPU-coated side to itself to create strong air-tight seals. The fabric was formed into the tubular shape of the vine robots by sealing two sheets together using a Vetron 5064 ultrasonic welder (Vetron Typical Europe GmbH, Kaiserslautern, Germany) for convenience, although they could also have been made using impulse sealers\cite{coad_vine_2020}. When inflated, the vine robots are 76.2 mm in diameter, and 119.7 mm wide when deflated and flattened. 

For the demonstrations using the small scale loop closure grasping system, we construct the vine robots using 0.15mm thick Low Density Polyethylene (ULINE, North America) that is 64.7 mm in diameter when inflated and 101.6 mm wide when deflated and flattened. The growth direction of vine robots can be actuated, generally, by four different methods: Distributed Strain, Concentrated Strain, Tip-Localized Strain and Preforming \cite{blumenschein_design_2020}. This work utilizes preforming for actuating the growth direction of the vine.

The base consists of a motorized winch inside of a pressurized chamber with an air-tight collar to fasten the base of the vine robot to, as illustrated in Fig. 3C. For the human lifting and long range demo, the pressurized chamber is made from aluminum to provide a strong grounding frame on which the winch can exert high forces onto the vine robot. The winch consists of a 31.8 mm diameter cylinder 3D printed from carbon-fiber infused nylon filament, with a 12.7 mm wide steel hex shaft core to withstand the high forces and torques applied onto the vine robot. The end of the vine robot inner material is fastened to the cylinder using cloth-backed adhesive tape (Gorilla Glue, Inc., Ohio, USA), and the winch was controlled such that the material is wrapped around by at least 2.5 rotations at all times to maintain a high capstan friction load capacity. For the fastening strength of the base, the point of connection is between the inner material of the vine robot and the motorized winch that winds/unwinds it. When lifting an object, the winch winds the inner material up while the outer material crumples its length, and thus the inner material bears the entire load. We leverage capstan friction to implement high-strength fastening by ensuring that the vine robot inner material was always wrapped around the winch over some minimum angle value so the holding force is always magnified. This amplification can be conservatively estimated using the Euler-Eytelwein formula: $T_{load}=T_{hold} e^{\mu \theta}$, where $T_{load}$ is the load capacity, $T_{hold}$ is the holding force at the tip of the vine robot inner material, $\mu$ is the coefficient of friction, and $\theta$ is the wrapping angle. The holding force implemented using cloth-backed adhesive tape, and the coefficient of friction between the uncoated ripstop fabric and the 3D printed winch material was experimentally measured to be 27.2 kg and 0.2, respectively. The base winch angle limit was set such that the vine robot inner material was wrapped by at least 16 radians (254 mm excess length needed for 31.8 mm diameter winch), yielding a load capacity of 667.3 kg. Thus, if the fastening strength to the base was the bottleneck for the load capacity of the system, given that the base and tip fastenings only bear half of the load each, an object weighing up to 1334.6 kg could be lifted. This estimate is conservative because it does not account for the increased holding force due to the outer layers applying pressure onto the inner layers, previously demonstrated in \cite{osele_tip-clutching_2024}. In future iterations, the vine robot inner material would be taped directly to the steel shaft without the 3D printed cylinder to reduce the moment arm from the tensile loads, thus reducing the torque on the winch. The shaft is driven by a Neo 550 brushless DC motor (Rev Robotics LLC, TX, USA) with a 100:1 ratio gearbox (AndyMark Inc., IN, USA) capable of providing a total peak stall torque of 97 Nm. The base of the vine robot’s outer material is fit over the outside of the 3D printed collar and fastened using a steel hose clamp and 3D printed clamp inserts. The pulling strength of both the base and the tip-fastening winch is dependent on the gear ratio of their transmission and the maximum radius of their winch, including the thickness of the wound material. The retraction mechanisms in both devices are just the winch winding up the length of the vine robot. Thus, for a given actuator with a known stall torque, the lifting capacity is dependent only on the torque amplification of the transmission and the moment arm of the winch (i.e. force = torque / radius). The base and tip-fastening winch use the same motor with a stall torque of 0.97 Nm, and have transmission ratios of 100:1 and 200:1, respectively. Given an infinite available vine robot length, the maximum radii that the base and tip devices can accommodate due to the added thickness of the vine robot wrapping around the winch are 60.3 mm and 96.3 mm, respectively. Thus, the lifting force capacities for the base and tip are 164 kg and 205 kg, enabling an object weighing up to 328 kg and 410 kg to be lifted, respectively. 

% For the long-range demonstration, a single vine robot mechanism was used with a 64.7 mm diameter vine robot made from LDPE, a motorized base, and the same tip-fastening winch device used in previous demonstrations. The pressurized base is made of acrylic. The winch consists of a 101.6 mm diameter cylinder 3D printed from Polylactic Acid (PLA) filament, with a 6.35 mm wide aluminum D-shaft core. The end of the vine robot inner material is fastened to the cylinder using high-strength braided rope. The shaft is driven by a 24V Compact Square-Face DC Gearmotor (McMaster-Carr).

Our implementation of the tip-fastening mechanism is a winch with an embedded clamping mechanism to clamp the tip of the vine robot and wind it up to pull on the payload, as illustrated in Fig. 3B. The body of the winch is constructed primarily from 3D printed parts, aluminum plates, and a 15.9 mm wide steel hex shaft, with 9.5 - 12.7 mm (0.375 - 0.5 in) diameter partially-threaded bolts used to reinforce the 3D printed parts in addition to fastening them together. The winch is also driven by a Neo 550 brushless DC motor with a 100:1 ratio gearbox, with an additional 2:1 sprocket-chain transmission. The embedded clamp mechanism utilizes the winch body as the jaws to maintain a compact body, in that a section of the winch can be separated from the main body to open up the clamp. When the clamp is closed, the cylindrical shape of the winch is restored. The clamp is opened and closed by two lead screw linear actuators (Eco-Worthy, China) with a 1500 N pulling capacity, respectively. The non-backdrivable nature allows the high clamping force to be passively maintained. The surfaces of the jaws were designed to maximize the friction on the vine robot by (1) lining them with a nonslip film (Dycem, RI, USA), and (2) implementing an interlocking wave surface pattern into the surfaces of the jaws. As discussed in “Component design” section, clamping the vine robot between the wave-patterned jaws wraps it around a series of curves (with equal radius and arc length) that amplifies the load capacity relative to the holding force (frictional clamping force without the wave pattern) through the capstan friction effect. The clamp was designed with n=8 equivalent curve segments with $\theta_c=\frac{pi}{2}$ (n reduced from actual value of n=9 to account for fillets reducing $\theta_c$ for the curves at the ends of the chain), and two non-backdrivable linear actuators that apply a combined clamping force of $F_{clamp}=299.4$ kg. The coefficient of friction between the clamp surface (lined with a high friction film) and the TPU-coated side of the ripstop fabric was experimentally measured to be 0.49. Thus, given equation 1, the theoretical load capacity is 149,574 kg. 

The pressure in the base is commanded using a QB4TANKKZP25PSG Electro-Pneumatic Pressure Regulator (Proportion-Air, Inc. IN, USA). All electronics are powered using a 110V AC to 12V DC AC-to-DC Power Converter (NUOFUWEITM Guangdong, China).

For the “grasping heavy yet fragile objects” demonstration in which a watermelon was grasped and lifted, a different, smaller grasping device designed within the same system framework was used. The bounding boxes of its bases are 4.35\% the size of the bases used in the human lifting demonstration by volume ($6.28\times10^{-4} m^3$ vs. $1.44\times10^{-2} m^3$), and the diameters of its vine robots are 32.9\% of those used in the human lifting demonstration (25 mm vs. 76 mm). Its bases and tip-fastening mechanisms are made primarily out of 3D printed Polylactic acid (PLA) instead of aluminum, and its vine robots are made from low-density polyethylene (LDPE) instead of nylon fabric, which has a lower but still sufficient tensile strength of 10 MPa \cite{szlachetka_low-density_2021}.  

For the demonstration of grasping a ball in a woven configuration, the designs of the vine robot, base, and tip-fastening mechanism were the same designs used in the watermelon grasping and lifting demonstration. Instead of using two vine robots parallel to each other, this system consisted of four 24.3 mm diameter vine robots that grow downward to the side of the object and then bend to grow under the object in a woven warp-weft pattern, each in a direction offset $90^{\circ}$ from the last, as shown in Fig.4A.

For the demonstration of the topologically interlocking grasp with the bucket handle, the large scale system  was used. In the “Grasp Versatility” section, we state that this interlocking grasp is theoretically infinitely stable, assuming no breaks in the loops. Notably, to uphold this assumption, the strength of the system can theoretically be scaled without affecting the grasp. For the connections with the ground, the fastening mechanism can be made theoretically infinitely strong, since the fastening strength does not impact the interaction between the grasping mechanism and the object. For the base connection, high-strength fastening is made trivial through the use of standard fastener hardware and/or adhesives. For the tip-fastening mechanism, high-strength holds can be achieved either with high-strength actuators, or better yet, with latching and clutching mechanisms that can powerfully hold the mechanism without needing powerful actuators \cite{osele_tip-clutching_2024}. Thus, assuming the base and tip grounding connections are sufficiently strong, the bottleneck for the grasping device’s pull-out force becomes the tensile yield force of the closed-loop mechanism, which can be made extremely high using high-strength fabrics such as nylon ripstop and/or reinforcing the structure with high-strength fibers such as Kevlar and carbon fiber threads/filaments.

\newpage
\subsection*{Supplementary Discussion S7. Human contact pressure experiment }

We measured the pressure distribution on the lower back of a 79.4kg (175lb) life-sized manikin (Simulaids I.A.F.F. Rescue Randy) lying in a supine position on a bed while a vine robot grows between its back and the bed to harness it. During the experiment, the vine robot grows downward out of its base until it reaches the mattress, at which point it is actuated to grow along the surface of the mattress toward and then underneath the manikin body without sliding friction. The manikin was outfitted with a custom soft capacitive sensor array created by Pressure Profile Systems (Los Angeles, CA, US) on its lower back as shown in Fig 5. The sensor array records pressure data in 6.45 x 10-4 m$^2$ cells at 20 Hz. The device has a range from 0 kPa to 20.48 kPa with 0.0276 ± 0.00689kPa measured resolution. It was calibrated linearity of 99.9 $\pm$ 0.1\% and signal-to-noise ratio of 732 ± 219. The sensor array covered a total area 0.0387 m2 (0.38m x 0.1m).

\newpage
\subsection*{Supplementary Discussion S8. Derivation of tip-fastening winch mechanism load capacity model}

The wave-patterned jaws and the belt (i.e. the deflated vine robot membrane) clamped between them are illustrated in Fig. S5. $\theta_i$ is the arc length of curve i, n is the number of curves in sequence, $T_{hold,i}=T_{h,i}$ and $T_{load,i}=T_{l,i}$ are the initial and amplified tensile load capacities of the belt before and after being wrapped around curve i, respectively, and $\mu$ is the coefficient of friction between the jaws and the belt material. The bend angle that the belt experiences when passing onto the first curve from the previous adjacent flat segment of the jaws, and when passing from the last curve onto its adjacent flat segment, are denoted as $\phi_h$ and $\phi_l$, respectively. Based on these definitions and given the Euler-Eytelwein formula: $T_l=T_h e^{\mu \theta}$, the holding and loading forces of each adjacent curve pair can be expressed as:
\begin{equation}
	T_{h,i+1} = T_{l,i} = T_{h,i} e^{\mu \theta_i}
\end{equation}
The amplified tensile load capacity of a curve i can then be expressed in terms of the holding force of the initial curve as: 
\begin{equation}
\begin{split}
    T_{l,i} = T_{h,i} e^{\mu \theta_i} &= T_{l,i-1} e^{\mu \theta_i} = T_{h,i-1} e^{\mu \theta_{i-1} } e^{\mu \theta_i} = T_{l,i-2} e^{\mu \theta_{i-1} } e^{\mu \theta_i } \\
    ... & = T_{h,1} e^{\mu \theta_1} ... e^{\mu \theta_{i-1}} e^{\mu \theta_i} =T_{h,1} e^{\mu \sum_1^i \theta_i }
\end{split}
\end{equation}
To include the amplifications of all of the curves as well as the bends in the belts at the transitions onto the flat segments at the ends of the chain of curves, as well as estimate the tensile holding force of the first curve as the unamplified frictional holding force from the clamping force $T_{h,1}=\mu F_{clamp}$, equation 2 can be expressed as: 
\begin{equation}
    T_{l,n}=T_{h,1} e^{\mu (\phi_h + \sum_1^n \theta_i + \phi_l)} = \mu F_{clamp} e^{\mu (\phi_h + \sum_1^n \theta_i + \phi_l)}   
\end{equation}
In the case of our tip-fastening winch design, $\theta_i$ is constant for all i ($\theta_i=\theta_c$), and $\phi_h = \phi_l = \frac{1}{2} \theta_c$. Thus, 
\begin{equation}
    T_{l,n} = \mu F_{clamp} e^{\mu(\frac{1}{2} \theta_c + n \theta_c + \frac{1}{2} \theta_c)} = \mu F_{clamp} e^{\mu (n+1) \theta_c}
\end{equation}
This estimate is also conservative because it does not account for the clamping force that is applied not just at the far end of the chain of curves (where $T_{hold}$ is defined), but also over its entire length. 

\begin{figure} % Do not use \begin{figure*}
	\centering
	\includegraphics[width=0.9\textwidth]{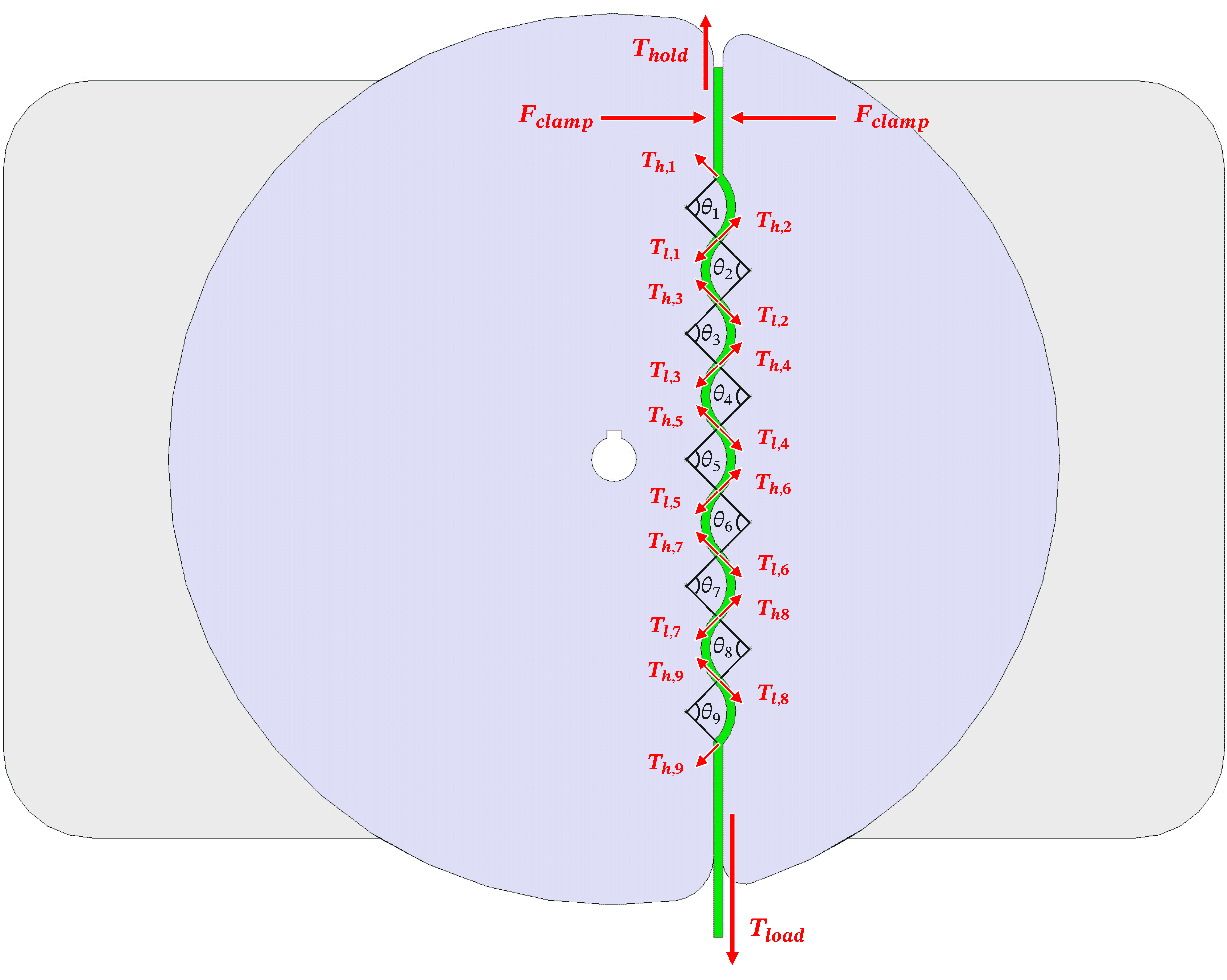} % for an image file named example_figure.*
	% Pick an appriopriate width for the size of the image

	% Captions go below figures
    {\linespread{1.0}
	\caption{\textbf{Wave pattern of jaws clamping deflated vine robot in tip-fastening winch mechanism.} Capstan friction between vine robot and curved segments of clamp jaws amplifies the clamping force to yield a high load capacity.}}
	\label{fig:TCW_holdForce} % give each figure a logical label name
\end{figure}

\newpage
\subsection*{Supplementary Discussion S9. Human grasping and lifting demonstration – participant interview responses}

Each of the following questions were verbally asked to the subject in person after they complete all experimental procedures.\\ \\
Question 1:	\quad Please rate the level of pain you experienced while being lifted by the robotic harness system on a scale of 1 to 7 (1 = none, 7 = unbearable).\\ \\
Response 1:	\quad 1 (“No pain.”)\\ \\
Question 2:	\quad Please rate the level of discomfort you experienced while being lifted by the robotic harness system on a scale of 1 to 7 (1 = none, 7 = unbearable).\\ \\
Response 2:	\quad 3 (“Holding my head up while the harness was lifting, produced a bit of strain on my neck and head.”)

\newpage
\subsubsection*{Description of Movies}

\textbf{Movie S1: Woven grasp configuration.} Small-scale loop closure grasping system grasps, lifts, and releases a ball with the linkages creating a woven caging configuration.
\\
\\
\textbf{Movie S2: Grasping via interlocking closed loops.} Small-scale loop closure grasping system creates and releases interlocked Hopf link grasps with a ring, and large-scale loop closure grasping system creates and releases interlocked Hopf link grasps with a bucket handle. 
\\
\\
\textbf{Movie S3: Grasping in a cluttered environment.} Small-scale loop closure grasping system grasps and lifts a kettle bell weight from inside a cluttered bin filled with plastic objects. The soft growing linkage mechanism grows through spaces smaller than its diameter to navigate through the objects into the desired grasping configuration. 
\\
\\
\textbf{Movie S4: Antagonistic insertion of linkages between a vase and its resting surface.} Small-scale loop closure grasping system grasping and lifting a glass vase from its resting surface with no preexisting gap. Three soft growing linkage mechanism grow through the interface between the vase and table from opposite directions to balance the horizontal pushing force that each exerts on the object, and navigate past the interface into the desired grasping configuration. 
\\
\\
\textbf{Movie S5: Long distance grasp.} Large-scale loop closure grasping system grasps and pulls boxes from 3 m away, leveraging the far-reaching deployability of vine robots from compact bases.
\\
\\
\textbf{Movie S6: Simultaneous multi-object grasping.} Small-scale loop closure grasping system grasps, transfers, and releases a pile of eight pipes in an single bundle. 
\\
\\
\textbf{Movie S7: "In-hand" manipulation.} Small-scale loop closure grasping system grasps, lifts, and rolls a cylinder in place mid-air by retracting and extending the two ends of the loops at equal and opposite speeds. 
\\
\\
\textbf{Movie S8: Grasping a human lying on a bed.} Large-scale loop closure grasping system grasps, lifts, and releases a human subject laying on a bed. 
\\
\\
\textbf{Movie S9: Eversion pressure under human experiment.} Contact pressure map over time between a life-sized weighted manikin and a the soft growing linkage mechanism of the large-scale system as it grows under their body.
\\
\\ 
\textbf{Movie S10: Grasping and lifting a watermelon.} Small-scale loop closure grasping system grasps, transfers, and releases a 5.9 kg watermelon.

\endgroup

\end{document}